\documentclass[letterpaper]{article} 
\usepackage{aaai24}  
\usepackage{times}  
\usepackage{helvet}  
\usepackage{courier}  
\usepackage[hyphens]{url}  
\usepackage{graphicx} 
\usepackage{natbib}  
\usepackage{caption} 
\usepackage{algorithm}
\usepackage{algorithmic}
\usepackage{epsfig}
\usepackage{graphicx}
\usepackage{amsmath} 
\usepackage{amssymb}
\usepackage{multirow}
\usepackage{booktabs}
\usepackage{caption} 
\usepackage{caption}
\usepackage{ctable} 
\usepackage{subcaption}
\usepackage{amsmath}
\usepackage[utf8]{inputenc}
\usepackage{graphicx}
\usepackage{newfloat}
\usepackage{listings}

\DeclareMathOperator*{\argmin}{arg\,min}
\DeclareCaptionStyle{ruled}{labelfont=normalfont,labelsep=colon,strut=off} 
\floatstyle{ruled}
\newfloat{listing}{tb}{lst}{}
\floatname{listing}{Listing}
\title{ShapeBoost: Boosting Human Shape Estimation with Part-Based Parameterization and Clothing-Preserving Augmentation
}
\author{
   Siyuan Bian\textsuperscript{\rm 1},
   Jiefeng Li\textsuperscript{\rm 1},
   Jiasheng Tang\textsuperscript{\rm 3,\rm 4},
   Cewu Lu\textsuperscript{\rm 1,\rm 2}
}
\affiliations{
    \textsuperscript{\rm 1}Department of Computer Science and Engineering, Shanghai Jiao Tong University, Shanghai, China\\
    \textsuperscript{\rm 2}MoE Key Lab of Artificial Intelligence, AI Institute, Shanghai Jiao Tong University, Shanghai, China\\
    \textsuperscript{\rm 3}DAMO Academy, Alibaba group, Hangzhou, China\\
    \textsuperscript{\rm 4}Hupan Lab, Hangzhou, China\\
    \{biansiyuan, ljf\_likit, lucewu\}@sjtu.edu.cn, jiasheng.tjs@alibaba-inc.com

    \vspace{10pt}{
        \centering
        \captionsetup{type=figure}
        \includegraphics[width=0.95\textwidth]{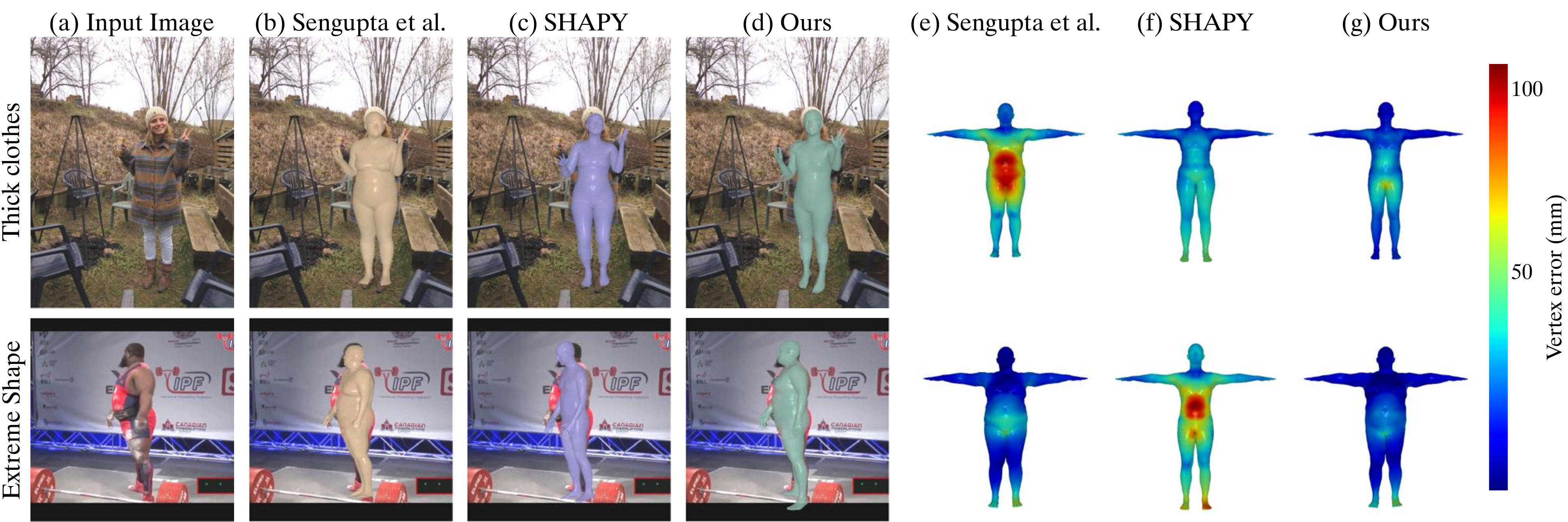}
        \vspace{-3pt}
        \captionof{figure}{Previous SOTA methods for human shape estimation~\cite{sengupta2021hierarchical,choutas2022accurate} (b, c) either fail on images of people wearing thick clothes or fail on images of people with extreme body shapes, while our method (d) achieves pixel-aligned results with high accuracy in both situations. Warmer colors on the human mesh represent higher per-vertex error.} 
        \label{fig:head_img}
    } \vspace{3.0pt}
}
\usepackage{bibentry}

\begin{document}
\maketitle

\begin{abstract}
    Accurate human shape recovery from a monocular RGB image is a challenging task because humans come in different shapes and sizes and wear different clothes. In this paper, we propose ShapeBoost, a new human shape recovery framework that achieves pixel-level alignment even for rare body shapes and high accuracy for people wearing different types of clothes. Unlike previous approaches that rely on the use of PCA-based shape coefficients, we adopt a new human shape parameterization that decomposes the human shape into bone lengths and the mean width of each part slice. This part-based parameterization technique achieves a balance between flexibility and validity using a semi-analytical shape reconstruction algorithm. Based on this new parameterization, a clothing-preserving data augmentation module is proposed to generate realistic images with diverse body shapes and accurate annotations. Experimental results show that our method outperforms other state-of-the-art methods in diverse body shape situations as well as in varied clothing situations. 
 \end{abstract}

 \section{Introduction}

 Human pose and shape (HPS) recovery from monocular RGB images is an essential task of computer vision. It serves as a basis for human behavior understanding and has applications in various fields such as Virtual Reality, Augmented Reality, and Autopilot. Recent methods~\cite{zhang2022pymafx,li2022cliff,li2022d,li2021hybrik} achieve high accuracy in human pose estimation, but their results of human shape estimation are often suboptimal.
 
 Due to the scarcity of image datasets featuring diverse body shapes, many existing methods for recovering human pose and shape suffer from overfitting on body shape estimation. Their results are particularly unsatisfactory for very thin or plump people. Previous approaches have attempted to solve the overfitting issue through two main strategies. The first kind of methods~\cite{varol2017learning,sengupta2020synthetic,sengupta2021probabilistic,sengupta2021hierarchical} train on synthetic data and exploit proxy representations to reduce the domain gap, while the second kind of methods~\cite{dwivedi2021learning,omran2018neural,agarwal2005recovering} exploit shape cues which are easy to annotate as weak supervision. However, for the first kind of methods, the synthetic images are unnatural with unrealistic texture and clothing, and the extracted proxy representations may be ambiguous and inaccurate. The situation is especially severe when the individual is wearing thick garments or is occluded in the image. For the second kind of methods, since 2D clues such as segmentations and silhouettes are highly correlated with the human pose and clothing, supervising with 2D clues may give wrong guidance of human shape in the case of inaccurate pose estimation or thick clothing. Moreover, the real-world images of extreme shapes are still insufficient. SHAPY~\cite{choutas2022accurate} improves the second kind of methods by using linguistic attributes and body measurements as supervision, which allows it making better estimates for clothed people. However, similar to other models trained on real-world datasets, it still performs poorly on images of people with extreme body shapes because of the lack of extreme body shapes in the training datasets. To sum up, just as shown in Fig.~\ref{fig:head_img}, the first kind of methods often fail on images with people in occlusion or thick clothing, while the second kind of methods often fail on images containing people with extreme body shapes. 
 
 To overcome the above limitations, we propose ShapeBoost, a new shape recovery framework based on a novel part-based shape parameterization. The new shape parameters are composed of bone lengths and mean widths of body part slices. Using a novel semi-analytical algorithm, the body shape can be accurately and robustly recovered from these parameters. During training, the bone lengths can be calculated from human keypoints, and the part widths are regressed by the neural network. Compared to the original shape parameters derived from PCA coefficients, our new part-based parameterization has a clear local semantic meaning, making it easier to regress and more flexible in application. During training, ShapeBoost augments new image-shape pairs by randomly transforming the raw image and calculating the corresponding part-based parameters. For image transformation, a clothing-preserving augmentation method is proposed: we first segment the human body out of the image and randomly transform it into a different shape. Then, the human segmentation is pasted back onto the inpainted background image with the guidance of the appearance consistency heatmap~\cite{fang2019instaboost}. The corresponding shape parameters can be analytically retrieved by applying the equivalent transformation since each component in the part-based representation is clearly defined.
 
 Compared to previous approaches, ShapeBoost generates realistic images of diverse human shapes in natural clothing together with the corresponding faithful annotations. Moreover, our new parameterization accurately describes the extreme body shapes and encourages pixel-level alignment. As a result, our method overcomes the disadvantages of existing methods and achieves high accuracy on images of people in thick clothes as well as on images of people with extreme body shapes. We benchmark our method on SSP-3D~\cite{sengupta2020synthetic} and HBW~\cite{choutas2022accurate} datasets. The results show that our method achieves state-of-the-art performance in both thick clothes situations and extreme body shape situations.

 The main contributions of this paper are summarized as follows:
 \begin{itemize}
    \item We present an accurate and robust human shape parameterization together with a semi-analytical shape recovery algorithm, which is flexible and interpretable.
    \item We propose ShapeBoost, a human shape recovery framework consisting of the a clothing-preserving data augmentation module and a shape reconstruction module.
    \item Our approach outperforms previous approaches and can handle diverse clothing as well as extreme body shapes. 
 \end{itemize}

\begin{figure*}[!ht]
    \begin{center}
    \includegraphics[width=0.9\linewidth]{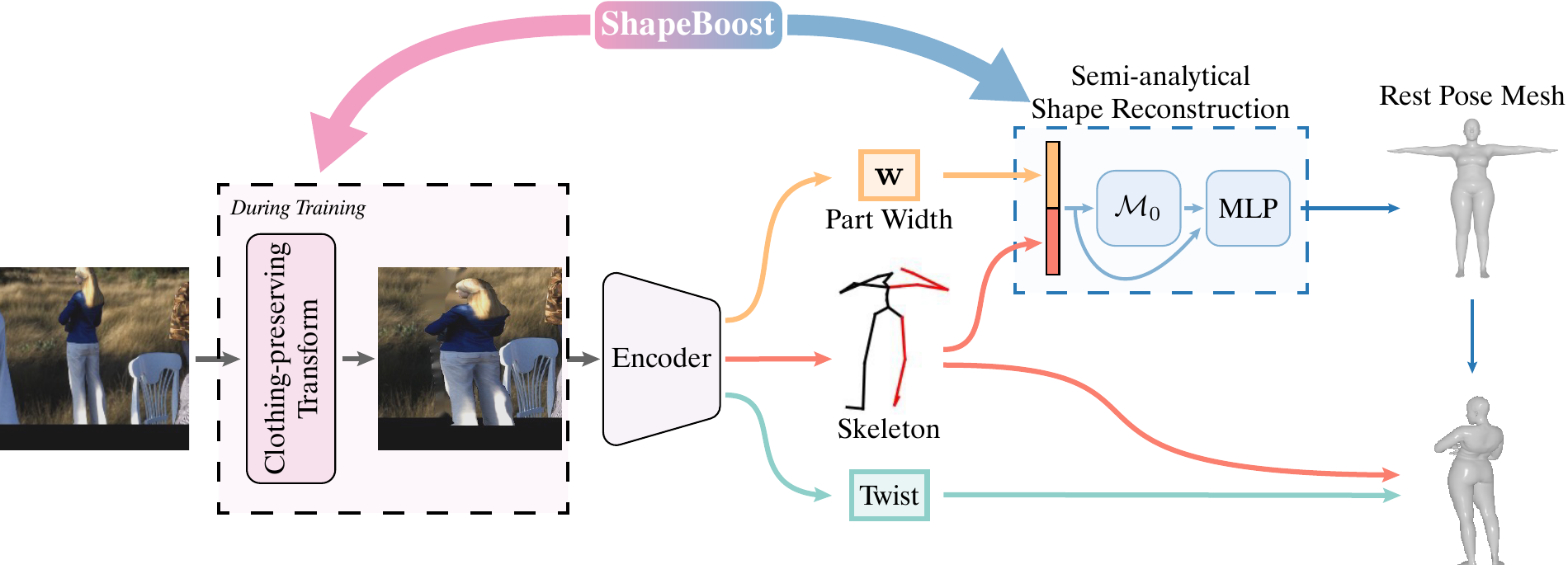}
    \end{center}
    \vspace{-2mm}
       \caption{
            {The overall pipeline}. 
            First, the input image is randomly transformed with the clothing-preserving image transformation, and a convolutional neural network (CNN) is employed to extract skeleton, part widths and twist rotations. Then, the pose is obtained using inverse kinematics and the shape is obtained with our semi-analytical algorithm. The final mesh is retrieved based on the pose and shape parameter. The ShapeBoost framework consists of the image augmentation module and the shape reconstruction module.
        }\vspace{-2mm}
    \label{fig:pipeline}
\end{figure*} 

\section{Related Work}

\subsection{3D Human Pose and Shape (HPS)}
Many algorithms have been proposed for reconstructing human pose and shape from RGB images, which are broadly categorized into two types. Firstly, \textbf{model-based methods} estimate parameters of a parameterized human model. Some methods~\cite{bogo2016keep,pavlakos2019expressive,guan2009estimating} estimate human pose and shape parameters by optimization. Regression-based methods~\cite{hmr,vibe,kocabas2021pare,li2022cliff,li2021hybrik}, on the contrary, employ neural networks to estimate the parameters. To reduce the difficulty of regression, many regression-based methods employ intermediate representations, including keypoints~\cite{hmr}, silhouettes~\cite{pavlakos2018learning}, segmentation~\cite{omran2018neural} and 2D/3D heatmaps~\cite{tung2017self}, keypoints~\cite{li2021hybrik,li2023hybrik,li2023niki} etc. Some approaches~\cite{spin,muller2021self,joo2021exemplar} combine optimization and regression. Secondly, \textbf{model-free methods} directly predict free-form representations of the human body, with the position of body model vertices predicted based on image features~\cite{corona2022learned,kolotouros2019convolutional,varol2018bodynet,lin2021end,lin2021mesh,moon2020i2l}, keypoints~\cite{choi2020pose2mesh}, or segmentations~\cite{varol2018bodynet}. These medthods mostly focus on human pose estimation and their results of human shape estimation are often unsatisfactory.

\textcolor{black}{Our work belongs to the model-based category, and we adopt inverse kinematics to estimate the human pose similar to HybrIK~\cite{li2021hybrik} for simplicity. However, instead of directly regressing the shape parameters, we employ a flexible and interpretable parameterization and a new shape reconstruction pipeline to achieve more accurate and robust shape estimation. Our method can also be easily applied to different pose estimation backbones.}

\subsection{Estimating 3D Body Shape}
Most recent HPS estimation methods excel in precise pose estimation but exhibit limitations in accurately estimating the real human body shape under clothing. Some methods have attempted to address this issue, and they mainly focus on novel training datasets and the estimation framework.

\paragraph{\textbf{Training datasets for human shape estimation.}} Accurately annotating body shapes from 2D human datasets~\cite{coco} is hard, and commmonly-used 3D human datasets~\cite{3dpw, h36m} contain limited number of people. To overcome this limitation, some researchers have created synthetic image datasets by rendering the mesh generated by parameterized human models~\cite{hoffmann2019learning,sengupta2020synthetic,varol2017learning,weitz2021infiniteform}. However, it is difficult to obtain images with natural clothing and realistic scenes using the naive rendering. Recently, more realistic synthetic datasets~\cite{bertiche2020cloth3d,pumarola20193dpeople,liang2019shape,patel2021agora,black2023bedlam} have been proposed, which contain people in different clothing with the help of human scans, simulation or deep generative networks. Choutas et al.~\cite{choutas2022accurate} have proposed the Model-Agency dataset, which uses images from model agency websites labeled with linguistic attributes and measurements. Although these new datasets contain more diverse body shapes, most datasets still lack people with extreme body shapes, and the authenticity of synthetic images remains insufficient.

\paragraph{\textbf{Estimation Framework.}} Several methods~\cite{sengupta2020synthetic,sengupta2021probabilistic,sengupta2021hierarchical} train the network directly on synthetic data. To reduce the domain gap, they use proxy representations (PRs) as input, such as part segmentation masks~\cite{varol2017learning}, silhouettes~\cite{sengupta2020synthetic,ruiz2022human}, Canny edge detection results~\cite{sengupta2021probabilistic,sengupta2021hierarchical} or 2D keypoint heatmaps~\cite{sengupta2020synthetic,sengupta2021probabilistic,sengupta2021hierarchical}. Other work~\cite{dwivedi2021learning,omran2018neural,agarwal2005recovering} uses real-world data for training and exploits 2D shape cues as supervision. Bodypart segmentation masks~\cite{dwivedi2021learning,omran2018neural} and silhouettes~\cite{agarwal2005recovering} are widely used among them. LVD~\cite{corona2022learned} learns the vertex descent direction based on image-aligned features, and SHAPY~\cite{choutas2022accurate} uses linguistic attributes and body measurements as supervision. 

Unlike previous work, our method generates images with diverse human body shapes without altering clothing, lighting, and background details. Therefore, the diversity is rich and the domain gap is small. Since our framework utilizes our new parameterization, there is no ambiguity even when the human is in thick clothing and our method will not enlarge error even when the pose estimation is inaccurate.

\section{Method}
In this section, we present our solution for human shape recovery (Fig.~\ref{fig:pipeline}). First, we give background knowledge of the parameterization of SMPL model in Sec.~\ref{sec:prel}. Considering its drawbacks, a flexible and interpretable part-based human shape parameterization is proposed in Sec.~\ref{sec:decomp}. Based on this new parameterization, in Sec.~\ref{sec:shapeboost}, we design a new human shape recovery framework called ShapeBoost. The training pipeline and loss functions are described in Sec.~\ref{sec:loss}.

\subsection{Preliminary}
\label{sec:prel}
\paragraph{SMPL Model.} In this work, SMPL model~\cite{loper2015smpl} is employed to represent human body pose and shape. SMPL provides a differentiable function $\mathcal{V}(\boldsymbol{\theta}, \boldsymbol{\beta})$ that maps pose parameters $\boldsymbol{\theta} \in \mathbb{R}^{3J}$ and shape parameters $\boldsymbol{\beta} \in \mathbb{R}^{10}$ to a human mesh $\mathbf{V}$, where $J$ is the number of joints. The pose parameters $\boldsymbol{\theta}$ represent the relative rotation of body joints, and the shape parameters $\boldsymbol{\beta}$ are coefficients of a PCA body shape basis. SMPL model is drived in two steps:
\begin{align}
    \mathbf{T} &= \mathcal{S}(\boldsymbol{\beta}), \\
    \mathbf{V} &= \mathcal{V}(\boldsymbol{\theta}, \boldsymbol{\beta}) = \mathcal{P}(\boldsymbol{\theta}, \mathcal{S}(\boldsymbol{\beta})).
\end{align}

First, a rest-pose mesh $\mathbf{T}$ is constructed using function $\mathcal{S}$. Second, the rest-pose mesh is driven to the target pose by function $\mathcal{P}$. The shape of the mesh is determined only by $\boldsymbol{\beta}$, and the posing procedure does not change the body shape. \textcolor{black}{Most current methods regress shape parameters $\boldsymbol{\beta}$ directly. However, since most available training datasets lack people with diverse body shapes, these methods often overfit and fail to generalize to unseen body shapes. }

\subsection{Part-based Parameterization}
\label{sec:decomp}
\textcolor{black}{In this work, we propose a novel parameterization of human shape using bone lengths and widths of part slices. Compared to the $\boldsymbol{\beta}$ representation which uses a global descriptor of the body shape, this new representation allocates shape descriptors to local body parts. This allows the network to learn from local image features and thus alleviates the overfitting problem. Furthermore, our parameterization is more flexible and interpretable, allowing compatibility with our data augmentation procedure discussed in Sec. \ref{sec:shapeboost}.}

In our parameterization, the SMPL mesh is divided into $J=24$ segments according to the linear blending weight, and each segment has a corresponding central bone ended with two joints. The distance of one vertex from its corresponding bone is called the ``width'' of this vertex for short. Each body part is further sliced into $n$ components along the bone, and the mean widths of the vertices in these $n$ slices are used to represent the thickness of that part. The segmenting and slicing technique is visually illustrated in Fig.~\ref{fig:shape_demp}. In this way, the formula of SMPL model is converted to:

\begin{align}
    \mathbf{T} &= \mathcal{M}(\mathbf{l}, \mathbf{w}), \\
    \mathbf{V} &= \mathcal{P}(\boldsymbol{\theta}, \mathcal{M}(\mathbf{l}, \mathbf{w})),
\end{align}
where $\mathbf{l}\in R^{J-1}$ represents the bone lengths of the body skeleton and $\mathbf{w}\in R^{nJ}$ represents the mean widths of all part slices. Under our new representation, the SMPL model first derives a rest-pose mesh using $\mathcal{M}(\mathbf{l}, \mathbf{w})$, and then uses function $\mathcal{P}$ to drive the mesh to the target pose just like the original SMPL model.


\begin{figure}[t]
    \centering
    \includegraphics[width=0.95\linewidth]{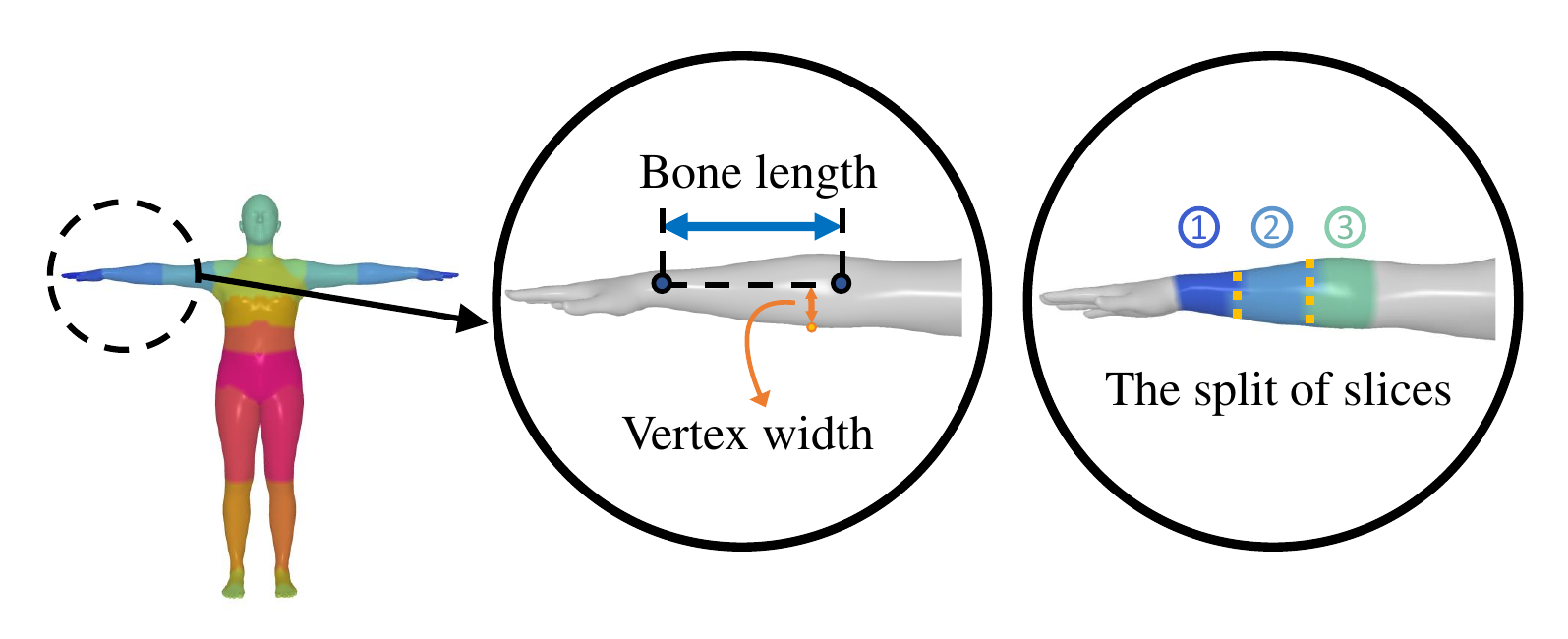} 

    \caption{{Illustration of the shape decomposition procedure.} From left to right, the figure shows the part segmentation, the definition of bone length and vertex width, and the slicing of one body part.}\vspace{-2mm}
    \label{fig:shape_demp}
\end{figure}

Deriving the function $\mathcal{M}$ directly by a neural network is untrivial and can lead to overfitting. Therefore, a semi-analytical algorithm is proposed that first solves a roughly correct mesh using analytical methods and then uses a multilayer perceptron (MLP) to correct the result using error feedback techniques. 

We can analytically retrieve a body shape that roughly conforms to the target bone lengths and part slice widths by (1) stretching the bones and broadening each part slice of the template mesh according to the target values. (2) using linear blend weights (LBS weights) to assemble these adjusted parts. \textcolor{black}{(3) using the PCA-coefficients of SMPL to retrieve the shape parameters from the deformed template mesh.} This mapping is referred to as $\mathcal{M}_0$. A detailed description of the analytical algorithm is available in supplementary materials. 

Since the input bone lengths and part widths often contain noise, the analytical algorithm sometimes produces suboptimal body shapes. Therefore, we use a 4-layer MLP to modify the analytically-retrieved shape parameters. The final formula of $\mathcal{M}$ can be written as 
\begin{equation}
    \mathbf{T} = \mathcal{M}(\mathbf{l}, \mathbf{w}) = \operatorname{MLP}(\mathcal{M}_0(\mathbf{l}, \mathbf{w}), \mathbf{l}, \mathbf{w}, \mathbf{\Delta l}, \mathbf{\Delta w}),~\label{eq:semi_analytical}
\end{equation}

where $\mathbf{\Delta l}$ and $\mathbf{\Delta w}$ are the difference between the target bone lengths and part slice widths and the corresponding values obtained by $\mathcal{M}_0$. In practice, instead of regressing the bone lengths directly, we extract the bone lengths from human keypoints. This setting further encourages the network to only focus on local, per-part image features and thus alleviate overfitting.

\subsection{ShapeBoost}
\label{sec:shapeboost}

Armed with the part-based parameterization discussed in Sec~\ref{sec:decomp}, we can manipulate the body shape in an intuitive way by stretching the bone lengths and broadening the part slice widths. These manipulations enable us to augment the raw human images and retrieve the new ground truth body shape which accurately explains the figure in the image after the transformation. This framework, named ShapeBoost, generates diverse body shapes while preserving clothing, lighting, and background details, and then takes use of our new parameterization to reconstruct the body shape. 

\paragraph{Clothing-preserving Image Transformation.}
An intuitive way to change the human shape in an image is to apply the affine transformation to the input image. For example, scaling an image with an aspect ratio unequal to 1 results in a visually thinner or ampler human figure. 

\begin{figure}[t]
    \centering
    \includegraphics[width=\linewidth]{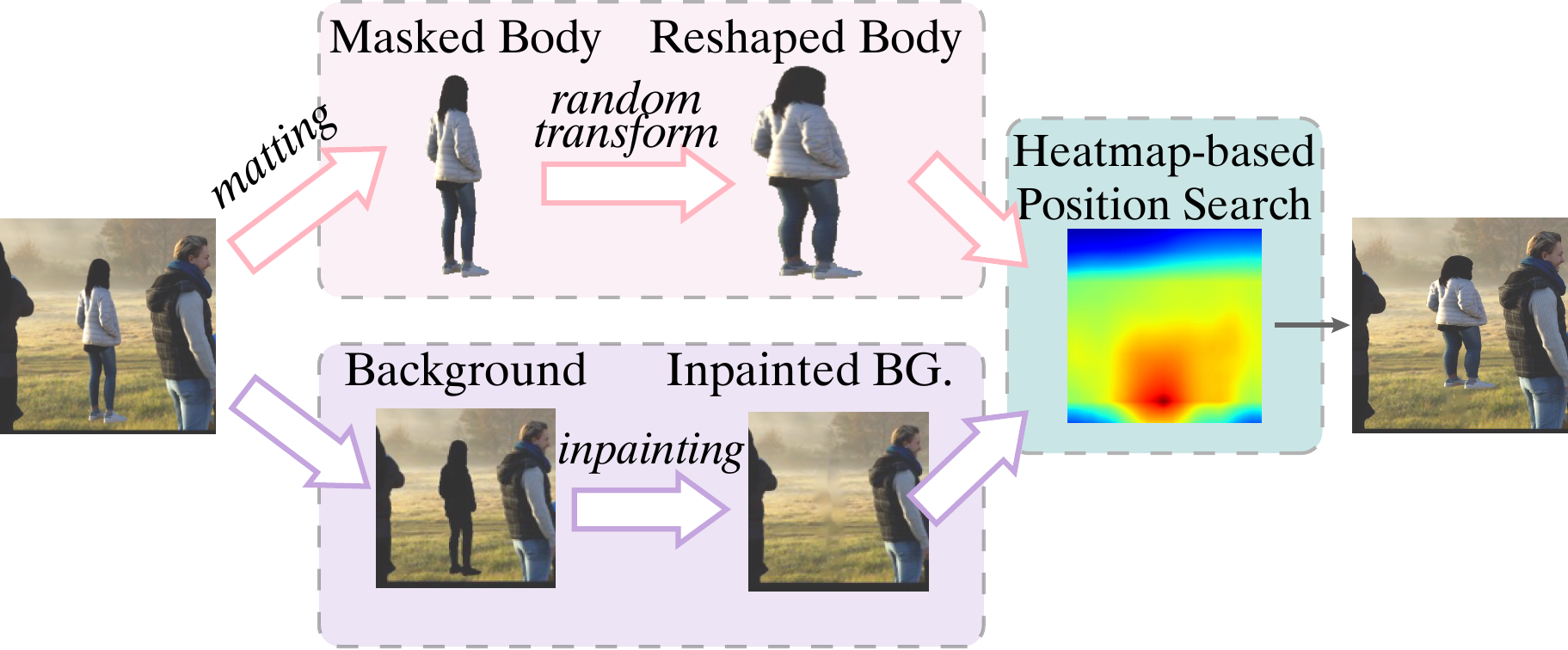}
    \caption{{The illustration of the clothing-preserving transformation.}}
    \label{fig:img_trans}\vspace{-2mm}
\end{figure}

However, applying the affine transform to the entire image results in a stretched background, which may leak the scaling information and thus incur overfitting. To alleviate this problem, we propose a silhouette-based augmentation method inspired by Instaboost~\cite{fang2019instaboost}. Instead of affine transforming the whole image, we first segment the human body out using the ground truth segmentation. Then we inpaint the background image, affine transform the segmented human body, and paste the transformed human body back onto the inpainted background image with the guidance of the appearance consistency heatmap~\cite{fang2019instaboost}. This method effectively avoids background stretching and produces more natural-looking images. The process is visually illustrated in Fig.~\ref{fig:img_trans}.

To simplify the discussion, we assume that the affine transformation consists of a rotation matrix and a scaling matrix, which is written as
\begin{equation}
    T = SR = 
    \left[
        \begin{array}{cc}
            a & 0 \\
            0 & b  \\
        \end{array}
    \right] 
    \left[
        \begin{array}{cc}
            \cos{\theta} & -\sin{\theta} \\
            \sin{\theta} & \cos{\theta} \\
        \end{array}
    \right] .
    \label{eq:affine}
\end{equation}

\paragraph{Shape-parameter Derivation.}
People in different poses are affected by the image transformation in different ways, which poses a great challenge for the derivation of the PCA-based shape parameters after the image transformation. However, with the part-based parameterization, we can still accurately explain the new body shape by estimating the widths and bone lengths of each body part. We use orthographic projection in our derivation.

Given the camera and pose parameters, the bone lengths after transformation can be easily obtained by stretching the bones to ensure a consistent 2D joint projection. Compared to the derivation of bone lengths, the derivation of the part slice widths after transformation is more complex. Suppose a vertex indexed by $k$ belongs to the $j$-th part. The distance of the vertex from the part bone on the 2D image plane, denoted by $w_{k}^{2D}$, is affected by the transformation according to the following equations:
\begin{equation}
    \bar{w}_{k}^{2D} = \frac{ab \cdot l_{j}^{2D}}{\bar{l}_{j}^{2D}} w_{k}^{2D},
    \label{eq:width_proj}
\end{equation}
where $l_{j}^{2D}$ and $\bar{l}_{j}^{2D}$ represent the bone lengths of part $j$ on the 2D image plane before and after the transformation, respectively; $a$ and $b$ are scaling factors mentioned in Eq.~\ref{eq:affine}. A detailed derivation is available in the supplementary materials. It is noteworthy that Eq.~\ref{eq:width_proj} implies the 2D widths of vertices on the same part are scaled by the same factor. Therefore, the underlining 3D part width of part $j$  is changed by
\begin{equation}
    \bar{w}_j = \frac{\bar{s}}{s}\times \frac{ab \cdot l_{j}^{2D}}{\bar{l}_{j}^{2D}}\times w_j .
\end{equation}\label{eq:w3d}
In the equation, $s$ and $w_j$ are the scale factor of the orthographic projection and the 3D part width of part $j$ before the image transformation, whereas $\bar{s}$ and $\bar{w}_j$ are the corresponding values after the transformation. Due to scale ambiguity, $\bar{s}$ is an ambiguous scaling factor that is difficult to directly derive. Therefore, in our training, we only supervise the projected results of the predicted part slice widths on the 2D image plane, without directly supervising their actual values. We hypothesize that the network can learn the best scaling factor $\bar{s}$ using the prior knowledge of human body shape.

\subsection{Training Pipeline and Loss Function}
\label{sec:loss}
The overall training pipeline is illustrated in Fig.~\ref{fig:pipeline}. First, the input image is transformed using the clothing-preserving image transformation, and the convolutional neural network (CNN) backbone is utilized to process the augmented image and estimate the skeleton (3D keypoints extracted from heatmaps), twist angles and part slice widths. \textcolor{black}{Second, we use these estimated values to reconstruct the pose and shape of the individual. The pose parameters are obtained with inverse kinematics similar to HybrIK~\cite{li2021hybrik}, while the shape parameters are retrieved using the semi-analytical algorithm discussed in Sec.~\ref{sec:decomp}. The final mesh is obtained based on the pose and refined shape parameters.} 

We employ end-to-end training for the pipeline, and the loss function consists of three components: shape loss, pose loss, and shape-decompose loss. The CNN backbone is supervised by shape loss and pose loss, while the MLP used in the shape reconstruction module is supervised by shape-decompose loss. 

\paragraph{Shape Loss.}
In shape loss, we supervise the predicted part widths predicted by the CNN backbone. Specifically, we require the projection results of the part slice widths and the vertex widths to be close to the target value after data augmentation. $K$ represents the number of vertices in the human mesh model and $J$ represents the number of joints.
\begin{equation}
    L_{shape} = \sum_j^J \| \hat{w}_j^{2D} - \bar{w}_j^{2D} \|_2^2 \ + \ \mu_0\sum_k^K \| \hat{w}_k^{2D} - \bar{w}_k^{2D} \|_2^2.
\end{equation}

\paragraph{Pose Loss.} Pose loss is designed to supervise the predicted skeleton and twist angle. We adopt the same loss function as HybrIK~\cite{li2021hybrik} and denote it as $L_{pose}$.

\paragraph{Shape-decompose Loss.} 
Shape-decompose loss ensures that the shape reconstruction module predicts a valid human mesh while best preserving the part slice widths and bone lengths predicted by the CNN backbone. It consists of three loss functions
\begin{equation}
    L_{decomp} = L_{bone} + L_{width} + \mu_1 L_{reg},
\end{equation}
where
\begin{align}
    &L_{bone} = \sum_j^J \left( \| \tilde{\mathbf{x}}_j - \hat{\mathbf{x}}_j \|_1 \ + \ \| \tilde{l}_j - \hat{l}_j\|_1 \right), \\
    &L_{width} = \sum_j^J \left( \| \tilde{w}_j - \hat{w}_j \|_2^2 \ + \ \| \frac{\tilde{w}_j}{\tilde{l}_j} - \frac{\hat{w}_j}{\hat{l}_j} \|_2^2 \right), \\
    &L_{reg} = \| \tilde{\boldsymbol{\beta}} \|_2^2.
\end{align}

In the equations, $\tilde{\mathbf{x}}_j$, $\tilde{l}_j$, $\tilde{w}_j$ are the keypoint coordinates, the bone length and the part slice widths of part $j$ refined by the MLP in the shape reconstruction module. $L_{bone}$ and $L_{width}$ supervise the preservation of the bone length and part slice widths respectively, and $L_{reg}$ regularizes $\boldsymbol{\tilde{\beta}}$ parameter. 

\paragraph{Overall Loss.} 
The overall loss of our pipeline is formulated as
\begin{equation}
    L = L_{pose} + \mu_2 L_{decomp} + \mu_3 L_{shape}.
\end{equation}

\section{Experiments}

\subsection{Datasets}
We use {3DPW}~\cite{3dpw}, {Human3.6M}~\cite{h36m}, {COCO}~\cite{coco}, {AGORA}~\cite{patel2021agora} and {Model Agency Dataset}~\cite{choutas2022accurate} for training. The original {Model Agency Dataset} contains $94,620$ images of $4,419$ models, but we only use about one-third of these images in our training due to the unavailability of many images on the Internet. To avoid data bias, the images are sampled following previous work~\cite{choutas2022accurate}. We also follow previous work and use synthetic data to assist network training. The rendering settings are identical to ~\cite{sengupta2021hierarchical}.

We evaluate our model on {SSP-3D}~\cite{sengupta2020synthetic} and {HBW datasets}~\cite{choutas2022accurate}. The results on SSP-3D dataset show the model's performance on diverse human body shapes, while the results on HBW dataset indicate the model's performance on images of people wearing different clothing.

\begin{table}
	\centering
	\resizebox{0.95\columnwidth}{!}
    {
        \begin{tabular}{l|l|l}
        \toprule
        Method & Model & PVE-T-SC $\downarrow$  \\
        \midrule
        HMR~\cite{hmr} & SMPL & $22.9$   \\
        SPIN~\cite{spin} & SMPL & $22.2$  \\
        (Sengupta et al. 2020) & SMPL & $15.9$ \\
        (Sengupta et al. 2021b) ${^\dagger}$ & SMPL & 13.3  \\
        (Sengupta et al. 2021a) & SMPL & 13.6 \\
        HybrIK~\cite{li2021hybrik} & SMPL & 22.8 \\
        LVD~\cite{corona2022learned} & SMPL & 26.1\\
        CLIFF~\cite{li2022cliff} & SMPL & 18.4 \\
        SHAPY~\cite{choutas2022accurate} & SMPL-X & 19.2 \\
        SoY~\cite{sarkar2023shape} & SMPL & 15.8\\
        \cite{ma20233d} & SMPL & 18.8 \\
        \midrule
        (Sengupta et al. 2021a)$^*$ & SMPL & 15.4 \\
        SHAPY~\cite{choutas2022accurate}$^*$ & SMPL &  12.2 \\
        \midrule
        ShapeBoost (Ours) & SMPL & \textbf{11.4} \\
        ShapeBoost (Ours) & SMPL-X & 12.0 \\
        \bottomrule
    \end{tabular}
    }
\caption{{Quantitative comparisons with state-of-the-art methods on the SSP-3D test set} in $\mathrm{mm}$. Symbol ${\dagger}$ means using multiple images as input, and symbol $*$ means retraining using the same training setting as our method.}\vspace{-2pt}
\label{table:ssp-3d}
\end{table}{}

\subsection{Comparison with the State-of-the-art}

\textcolor{black}{We evaluate the performance of different methods on SSP-3D and HBW test and validation datasets. Following previous work, on SSP-3D dataset, we use PVE-T-SC, a scale-normalized per-vertex error metric to evaluate the model performance. On HBW dataset, we report the predicted height (H), chest (C) , waist (W), and hip circumference (HC) errors, and P2P$_{20\mathrm{K}}$ errors of different models. All the experiments of our method use part slicing number $n=1$ by default unless otherwise stated. For a fair comparison, we also retrain two best-performing networks~\cite{sengupta2021hierarchical,choutas2022accurate} with the same datasets and settings as our method.} 

\begin{figure*}[t]
  \begin{center}
  \includegraphics[width=0.95\linewidth]{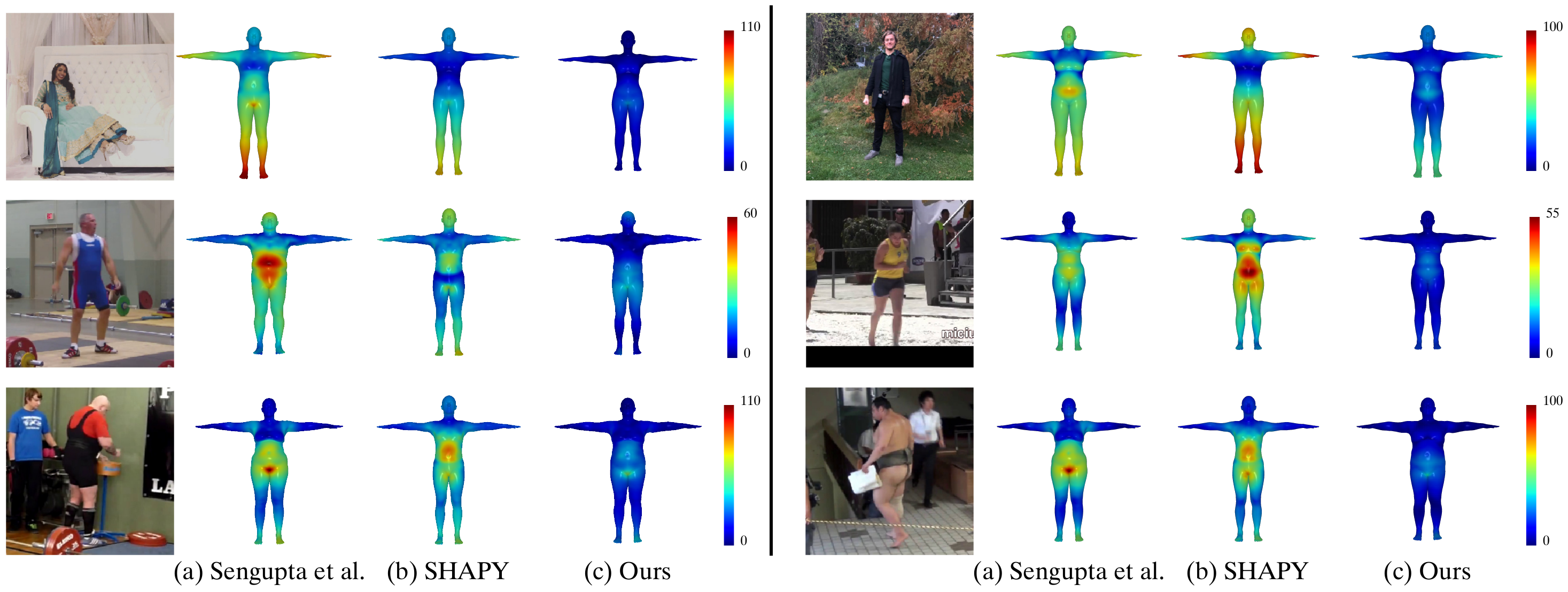}
  \end{center}
  \vspace{-2mm}
     \caption{
      {Qualitative results on SSP-3D and HBW datasets.} From left to right: Input image, (a) Sengupta et al.~\cite{sengupta2021hierarchical} results, (b) SHAPY~\cite{choutas2022accurate} results, and (c) Our results. Warmer colors mean higher per-vertex error. Experiments on SSP-3D dataset use PVE-T-SC metric, and experiments on HBW dataset use P2P$_{20\mathrm{K}}$ metric.
      }
  \label{fig:qualitative} 
\end{figure*}

\textcolor{black}{Tab.~\ref{table:ssp-3d} shows that our method surpasses previous works on SSP-3D dataset, which shows that our method can deal with the diverse human body shape much better than previous methods. Tab.~\ref{table:hbw-val} and ~\ref{table:hbw-test} shows the performance on HBW validation and test dataset. On HBW test dataset, our method achieves comparable results with previous SOTA methods and predicts more accurate waist and hip circumferences. On HBW validation set, our method outperforms previous SOTA methods. These results prove that our method can deal with diverse human clothing better than previous methods.  Qualitative results are provided in Fig.~\ref{fig:qualitative}.}

\begin{table}
	\centering
	\resizebox{\columnwidth}{!}
    {
    \begin{tabular}{l|lllll}
        \toprule
        Method & {H} & {C} & {W} & {HC} & P2P$_{20\mathrm{K}}$ \\
        \midrule
        SPIN & 59 & 92 & 78 & 101 & 29 \\
        Sengupta et al. 2020  & 135 & 167 & 145 & 102 & 47 \\
        TUCH & 58 & 89 & 75 & 57 & 26 \\
        Sengupta et al. 2021a & 82 & 133 & 107 & 63 & 32 \\
        CLIFF &  - & - & - & - & 27 \\
        SHAPY  & \textbf{51} & 65 & 69 & 57 & \textbf{21} \\
        \midrule
        ShapeBoost (SMPL)  & 66 & \textbf{63} & 58 & \textbf{47} & 25 \\
        ShapeBoost (SMPL-X) & 68 & 69 & \textbf{56} & 49 & 22 \\
        \bottomrule
    \end{tabular}
    }
\caption{{Quantitative comparisons with state-of-the-art methods on the HBW test set} in $\mathrm{mm}$.}
\label{table:hbw-test}
\end{table}{}


\begin{table}
	\centering
	\resizebox{0.97\columnwidth}{!}
    {
        \begin{tabular}{l|lllll}
        \toprule
        Method & H & C & W & HC & P2P$_{20\mathrm{K}}$ \\
        \midrule
        Sengupta et al. 2021a  & 68 & 89 & 111 & 71 & 30 \\
        HybrIK  & 88 & 82 & 74 & 51 & 33\\
        LVD $^\#$  & - & 89 & 131 & 87 & 31\\
        SHAPY & 63 & 59 & 85 & 54 & 25 \\
        Ma et al. 2023 & 112 & 87 & 133 & 59 & 41 \\
        \midrule
        Sengupta et al. 2021a$^*$ & 72 & 66 & 74 & 49 & 29 \\
        SHAPY$^*$ & 62 & 52 & 72 & 50 & 26 \\
        \midrule
        ShapeBoost (SMPL) & \textbf{58} & 54 & 72 & \textbf{42} & 25 \\
        ShapeBoost (SMPL-X) & 61 & \textbf{49} & \textbf{71} & 49 & \textbf{23} \\
        \bottomrule
    \end{tabular}
    }
\caption{{Quantitative comparisons with state-of-the-art methods on the HBW validation set} in $\mathrm{mm}$. Symbol $\#$ means using ground truth scale and symbol $*$ means retraining using the same training setting as our method.} 
\label{table:hbw-val}
\end{table}{}
 
\subsection{Ablation Study}
To demonstrate the effectiveness of different components in our method, we conduct ablation studies on SSP-3D dataset and HBW validation set. 

\paragraph{Shape reconstruction.}
To analyze the effectiveness and robustness of our new human shape parameterization, we reconstruct body shapes using bone lengths and part slice widths with different reconstruction algorithms under different noise ratios. The results are shown in Tab.~\ref{table:ablation_n}. All the model are trained on shape parameters sampled from Gaussian distributions and tested on 500 different body shapes obtained from AMASS dataset~\cite{mahmood2019amass}. ``Hybrid'' algorithm means using the semi-analytical algorithm, ``Analytical'' algorithm means solely employing the analytical algorithm, and ``NN'' algorithm means directly using the neural network without analytical steps.  From the first three lines in Tab.~\ref{table:ablation_n}, we observe that our proposed semi-analytical algorithm achieves the lowest error especially when the noise ratio is small. Additionally, when the noise is subtle, the parameterizations using different part slicing number ($n=1,2,3$) all achieve an acceptable low error. When the noise ratio is large, the error ratio decreases with larger $n$. Thus, we can conclude that our semi-analytically method accurately reconstructs human shape, and a larger $n$ makes it more robust to noise.

\paragraph{Shape estimation from images.}
We also experiment using different parameterizations for estimating human body shapes from RGB images. Tab.~\ref{table:ablation_n_img} provides a comparison of the results obtained using the direct shape parameterization ($\boldsymbol{\beta}$)~\cite{li2021hybrik} with our novel parameterization utilizing $n=1$ and $n=2$. We use image augmentation in the training. Since it is hard to find a ground truth $\boldsymbol{\beta}$ for augmented images, we use the 2D coordinates of vertices as supervision. We find that using our new parameterization yields better results, but a larger $n$ does not improve performance. The reasons are (1) the parameterization with $n=1$ already achieves a small shape reconstruction error (2) using larger $n$ complicates the regression task for the CNN backbone,  resulting in a reduction in the accuracy of predicting part slicing widths.

\paragraph{The effectiveness of data augmentation.}
We also make ablation studies with different training data quantitatively. The results are shown in Tab.~\ref{table:ablation_aug}. When the data augmentation module is not used, the performance of our model drops on both HBW and SSP-3D dataset. This shows the effectiveness of our data augmentation module.
\begin{table}
	\centering
	\resizebox{\columnwidth}{!}
    {
	\begin{tabular}{l|l|cccc}
		\toprule 
		\multicolumn{2}{c}{} & \multicolumn{4}{|c}{ V2V Error $(\mathrm{mm}) \downarrow$} \\
		\midrule
		$\textbf{n}$ & \textbf{Algo.} &\textbf{0\%noise} &\textbf{1\%noise} &\textbf{2\%noise} &\textbf{5\%noise} \\
		\midrule
		$1$ & Hybrid & 0.69 & 2.30 & 5.95 & 8.83  \\
		$1$ & Analy. & 6.14 & 6.59 & 8.99 & 12.34  \\
		$1$ & NN & 1.82 & 2.99 & 6.20 & 8.98  \\
		\midrule
		$2$ & Hybrid & \textbf{0.58} & 2.01 & 5.40 & 8.21 \\
		$3$ & Hybrid & 0.65 & \textbf{1.93} & \textbf{5.00} & \textbf{7.63}   \\
		\bottomrule
	\end{tabular}
	}
	\caption{{Ablation experiments of reconstructing shape using our new shape parameterization} in $\mathrm{mm}$.}\vspace{-2pt}
	\label{table:ablation_n}
\end{table}{}
\begin{table}
	\centering
	\resizebox{0.6\columnwidth}{!}
    {
    \begin{tabular}{l|ll}
        \toprule  Method & PVE-T-SC & P2P$_{20\mathrm{K}}$ \\
        \midrule
        $\beta$ & 12.3 & 26.0  \\
        $n=1$ &  \textbf{11.4} & \textbf{25.1}  \\
        $n=2$  &  11.6 & 26.2 \\
        \bottomrule
    \end{tabular}
    }
\caption{{Ablation experiments of shape estimation from RGB images} using different shape parameterization on SSP-3D and HBW validation set in $\mathrm{mm}$.}
\label{table:ablation_n_img}
\end{table}{}
\begin{table}
	\centering
	\resizebox{\columnwidth}{!}
    {
    \begin{tabular}{l|ll}
        \toprule  
        Method & PVE-T-SC & P2P$_{20\mathrm{K}}$ \\
        \midrule
        ShapeBoost (Ours) & \textbf{11.4} & \textbf{25.1}  \\
        w/o Augment & 12.1 & 26.5  \\
        w/o Augment, w/o Decompose & 12.4 & 27.0  \\
        \bottomrule
    \end{tabular}
    }
\caption{{Ablation experiments of data augmentation module} on SSP-3D and HBW validation set in $\mathrm{mm}$.}
\label{table:ablation_aug}
\end{table}{}

\section{Conclusion}
In this paper, we present ShapeBoost, a new framework for accurate human shape recovery that outperforms the current state-of-the-art methods. This framework exploits a new human shape parameterization that decomposes human shape into bone lengths and the mean width of each part slice. Compared to the existing representation with PCA coefficients, our new method is more flexible and interpretable. Based on the new shape parameterization, a new clothing-preserving data augmentation module is proposed to generate realistic images of various human shapes and the corresponding accurate annotations. Our method randomly augments the body shape without destructing the clothing details. Experiments show that our method achieves SOTA performance for extreme body shapes as well as achieves high accuracy for people under different types of clothing. 

~\newpage
\section{Acknowledgments} 
Cewu Lu is the corresponding author. He is the member of Qing Yuan Research Institute, Qi Zhi Institute and MoE Key Lab of Artificial Intelligence, AI Institute, Shanghai Jiao Tong University, China.

This work was supported by the National Key R\&D Program of China (No. 2021ZD0110704), Shanghai Municipal Science and Technology Major Project (2021SHZDZX0102), Shanghai Qi Zhi Institute, and Shanghai Science and Technology Commission (21511101200).

~\newpage

~\newpage

\appendix
{\bf{\huge Appendix}}
\vspace{2ex}%
 
In the supplemental document, we provide:
\begin{itemize}
   \item [Sec.~\ref{sec:param_detail}] Details of the proposed part-based parameterization.
   \item [Sec.~\ref{sec:param_der}] Details of ShapeBoost.
   \item [Sec.~\ref{sec:implementation}] Additional implementation details.
   \item [Sec.~\ref{sec:exp}] Additional experimental results.
   \item [Sec.~\ref{sec:smpl2smplx}] The method for converting the SMPL mesh to the SMPL-X mesh.
   \item [Sec.~\ref{sec:limitation}] Limitations and future work.
   \item [Sec.~\ref{sec:vis}] More qualitative results.
\end{itemize}
\section{Details of Part-based Parameterization}
\label{sec:param_detail}
In our part-based parameterization, we use a semi-analytical algorithm ($\mathcal{M}$) to reconstruct the human shape. Given the bone lengths and part slice widths, we first use an analytical algorithm $\mathcal{M}_0$ to obtain a rough mesh, and then use a multilayer perceptron to refine the mesh. In this section, we give the details of the analytical algorithm and provide an error analysis of $\mathcal{M}_0$ and $\mathcal{M}$.

\subsection{Details of $\mathcal{M}_0$}
In our parameterization, the SMPL mesh~\cite{loper2015smpl} is divided into $J=24$ segments according to the linear blending weight. Each vertex belongs to the body part which has the largest blending weight among all the joints. This splitting method is the same as that used in PARE~\cite{kocabas2021pare}. To analytically obtain a mesh whose bone lengths and part slice widths approximate the target values, the SMPL mean-shape template is modified according to (1) the target bone lengths, denoted as $\mathbf{l}$ and (2) the target part slice widths, denoted as $\mathbf{w}$. The bone lengths $\mathbf{l}$ are composed of the lengths of all pairs of joints connected in the kinematic tree. The part slice widths $\mathbf{w}$ consist of the mean width of each slice in different body parts. Suppose the part slicing number is $n$, and the number of body parts is $J$. The slice widths on the $j$-th part is denoted as $w_j$, $w_j = \left\{w_{j,1}, w_{j,2}, \dots w_{j,n}\right\}$, and on the whole body, the part slice widths $\mathbf{w} = \left\{ w_1, w_2, \dots, w_J\right\}$.

First, the rest-pose skeleton of SMPL template is stretched to ensure that the bone lengths match the target values. Suppose the bone with index $j$ connects two joints with index $j1$ and $j2$. The coordinates of these joints in the template mesh is denoted as $\mathbf{t}_{j1}$ and $\mathbf{t}_{j2}$, and the coordinates of these joints after stretching is denoted as  $\mathbf{x}_{j1}$ and $\mathbf{x}_{j2}$. $l_j$ represents the target lengths of this bone. These values satisfy the following equation:
\begin{equation}
    \mathbf{x}_{j2} = \mathbf{x}_{j1} + l_j \frac{\mathbf{t}_{j2} - \mathbf{t}_{j1}}{\| \mathbf{t}_{j2} - \mathbf{t}_{j1} \|_2}.
\end{equation}
This equation stretches the bone lengths to the target value while keeping the direction of each bone unchanged.

Second, the vertex positions on each part are adjusted according to $\mathbf{w}$. We split each human part into $n$ slices. The target body slice widths on part $j$ are denoted by $w_j = \left\{w_{j,1}, w_{j,2}, \dots w_{j,n}\right\}$, and the corresponding slice widths in the mean-shape template mesh are $v_j = \left\{v_{j,1}, v_{j,2}, \dots v_{j,n}\right\}$. Assume $\mathbf{p}_{k}$ is a vertex on the SMPL template mesh belonging to the $i$-th slice in part $j$. $\mathbf{q}_k$ is the projection point of $\mathbf{p}_k$ on the template's bone ended with $\mathbf{t}_{j1}$ and $\mathbf{t}_{j2}$. The new vertex position after the adjustment is computed as:
\begin{align}
    \mathbf{p}_{jk}^{\prime} = \mathbf{x}_{j2} &+ \frac{\left\| \mathbf{t}_{j1} - \mathbf{q}_k \right\|_2 }{\left\| \mathbf{t}_{j1} - \mathbf{t}_{j2} \right\|_2} (\mathbf{x}_{j1} - \mathbf{x}_{j2}) \\
    &+ \frac{ w_{j,i} }{ v_{j,i} } (\mathbf{p}_k - \mathbf{q}_k).
    \label{eq:decomp1}
\end{align}

This equation broadens the distance of each vertex from the bone while keeping its relative projection position along the bone unchanged.

The final coordinate of each vertex is linearly blended with each part: 
\begin{equation}
    \mathbf{p}_k^{\prime} = \sum_{j=1}^{J} w^{\textrm{lbs}}_{jk}\ \mathbf{p}_{jk}^{\prime} \ ,
    \label{eq:decomp2}
\end{equation}
where $w^{\textrm{lbs}}_{jk}$ is the LBS weight of the $k$-th vertex on part $j$.

\begin{figure}[t]
    \centering
    \includegraphics[width=\linewidth]{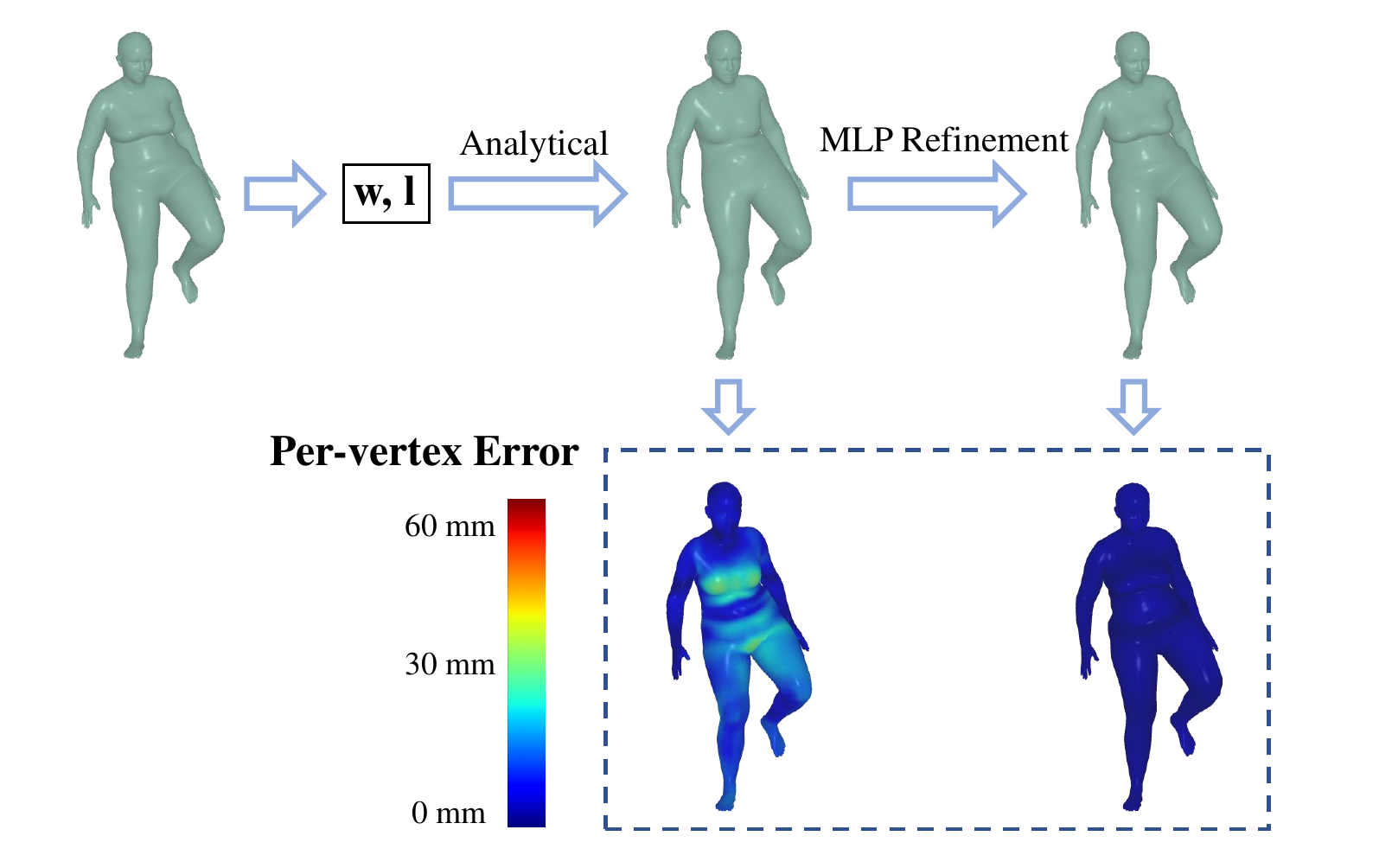} 

    \caption{\textbf{Error analysis of the shape recovery algorithms} used in the part-based parameterization. Given a ground truth mesh (top left), we first extract its bone lengths and part slice widths. Then, we reconstruct the human mesh with these extracted values using the analytical algorithm ($\mathcal{M}_0$) and MLP refinement ($\mathcal{M}$).}
    \label{fig:shape_demp} 
\end{figure}
In this way, the approximated SMPL mesh is analytically obtained. This mapping is referred to as $\mathcal{M}_0$. 

\subsection{Error Analysis of $\mathcal{M}_0$ and $\mathcal{M}$}
We also provide the error analysis of $\mathcal{M}_0$ and $\mathcal{M}$ for shape reconstruction in Fig.~\ref{fig:shape_demp}. Given a ground truth mesh, we first extract its bone lengths and part slice widths. Then, the human mesh is reconstructed with these extracted values using the analytical algorithm ($\mathcal{M}_0$) and MLP refinement ($\mathcal{M}$) in turn. From the per-vertex error heatmap, we can find that using the analytical algorithm ($\mathcal{M}_0$) alone already produces a mesh that is close to the ground truth mesh. However, since $\mathcal{M}_0$ generates the mesh by trivially stretching the template mesh, some details of the human form are lost. For example, the shape of the breasts and hips are slightly changed. The multilayer perceptron is then used to refine the mesh generated by $\mathcal{M}_0$. After the refinement, the details are recovered, and the per-vertex error drops to almost zero.




\section{Details of ShapeBoost}
\label{sec:param_der}
In this section, we provide some details of the shape parameter derivation, and give qualitative results of our data augmentation module. 

\subsection{Details of Shape-parameter Derivation}
Using the same notation as the main paper, we assume that the affine transformation consists of a rotation matrix and a scaling matrix. The transformation matrix is written as
\begin{equation}
    T = SR = 
    \left[
        \begin{array}{cc}
            a & 0 \\
            0 & b  \\
        \end{array}
    \right] 
    \left[
        \begin{array}{cc}
            \cos{\theta} & -\sin{\theta} \\
            \sin{\theta} & \cos{\theta} \\
        \end{array}
    \right] .
    \label{eq:affine}
\end{equation}

After applying the affine transformation to the image, the 2D bone length of the $j$-th part in the image plane ($\bar{l}_j^{2D}$) is changed by:
\begin{equation}
    \bar{l}_j^{2D} = \left\|\bar{\mathbf{x}}_{j1}^{2D} - \bar{\mathbf{x}}_{j2}^{2D}\right\|_2 = \left\|T(\mathbf{x}_{j1}^{2D} - \mathbf{x}_{j2}^{2D})\right\|_2,
\end{equation}
where $\mathbf{x}_{j1}^{2D}$ and $\mathbf{x}_{j2}^{2D}$ are coordinates of the bones's endpoints in the original image, and $\bar{\mathbf{x}}_{j1}^{2D}$ and $\bar{\mathbf{x}}_{j2}^{2D}$ are those coordinates in the new image after transformation.

Suppose a vertex indexed by $k$ belongs to the $j$-th part. On the 2D image plane, the equation always holds whatever $\theta$ is.
\begin{equation}
    \bar{w}_{k}^{2D}\bar{l}_{j}^{2D} = ab \cdot w_{k}^{2D}l_{j}^{2D},
\end{equation}
where $l_{j}^{2D}$ and $w_{k}^{2D}$ are the bone length of the $j$-th body part and the distance of the $k$-th vertex to the bone on the 2D image plane before transformation. $\bar{w}_{k}^{2D}$ and $\bar{l}_{j}^{2D}$ are the corresponding values after the transformation. 


Then we can get the 2D width of the vertex indexed by $k$ after the image transformation by:
\begin{equation}
    \bar{w}_{k}^{2D} = \frac{ab \cdot l_{j}^{2D}}{\bar{l}_{j}^{2D}} w_{k}^{2D}.
\end{equation} 

\subsection{Visualizing the Image Augmentation}
\begin{figure*}[ht]
    \centering
    \includegraphics[width=\linewidth]{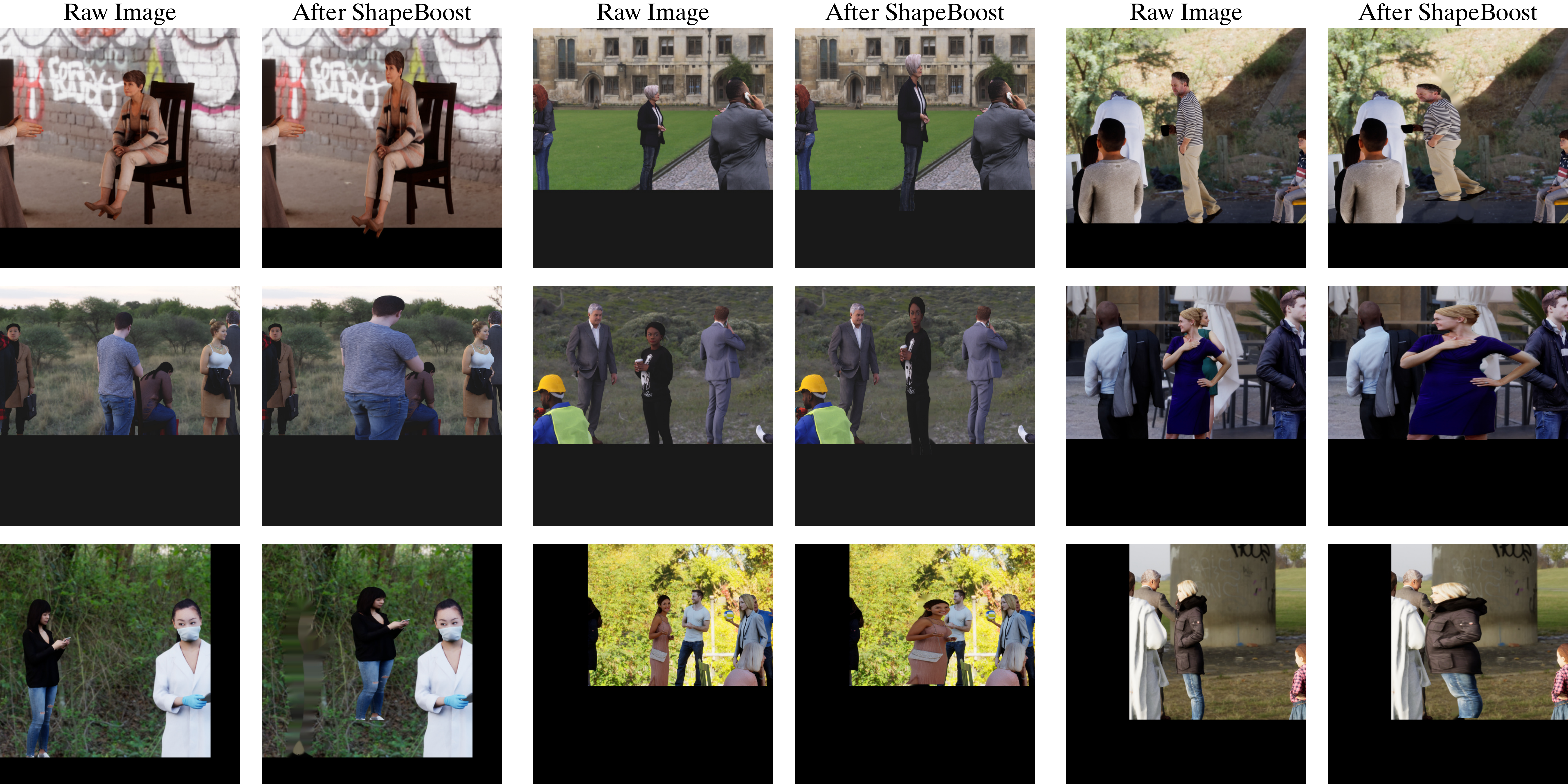} 

    \caption{\textbf{Qualitative results of images generated by the data augmentation module.} The generated images are realistic and include diverse body shapes.}
    \label{fig:aug_compare} 
\end{figure*}

We visualize the results of the clothing-preserving image augmentation in Fig.~\ref{fig:aug_compare}. We can find that our data augmentation module provides realistic images with diverse body shapes and natural clothing.

\section{Implementation Details}
\label{sec:implementation}
\paragraph{Detailed Model Structure} Following previous methods, we use Hrnet-W48~\cite{wang2020deep} as our backbone. The CNN backbone is initialized using pretrained weights from HybrIK~\cite{li2021hybrik}. The output of the CNN backbone is the 3D skeleton, the part slice widths and the twist angle. \textcolor{black}{After the backbone, the semi-analytical algorithm is used to refine the human body shape. The MLP used in the algorithm comprises 4 linear layers with LeakyReLU activation and hidden sizes of 512. It is pretrained on AMASS, and then trained end-to-end with the whole network. Following Eq. 5 in the main paper (in Sec. 3.2 in the main paper), the input of the MLP is the concatenation of (1) analytically-retrieved 10-dim shape, (2) the predicted body part widths regressed by the CNN, (3) the  bone lengths extracted from the predicted human keypoints, (4) the difference of bone lengths and part slice widths between the target values and the values obtained by the analytical algorithm $\mathcal{M}_0$. The output of the MLP is the refined 10-dim body shape.}  All the experiments use $n=1$ by default unless otherwise stated.

\paragraph{Datasets} We randomly apply data augmentation to $67\%$ of the input images. We utilize the silhouette-based augmentation paradigm when the subject is not occluded and the ground truth segmentation is available. For other cases, we use naive augmentation that affine transforms the entire image. The aspect ratio of the scaling factor in the affine transformation ($\frac{a}{b}$) is uniformly sampled from $0.4$ to $1.0$ with a probability of $33\%$, and uniformly sampled from $1.0$ to $2.5$ with a probability of $67\%$. This selection of probability aims to generate more images of people with chubby body shapes that the original datasets lack. After the affine transformation, the size of the bounding box is adjusted so that the subject is positioned in the center of the image and occupies a relatively large space. 

Following SHAPY~\cite{choutas2022accurate}, we use Model Agency Dataset~\cite{choutas2022accurate} in our training and utilize height, weight, chest/waist/hip circumference and linguistic shape attributes as weak supervision. After adding Model Agency Dataset, the model's performance on HBW validation set is slightly improved (P2P$_{20\mathrm{K}}$ drops by less than 1 point). We follow previous work and use fixed data sampling ratios for training. The sampling ratios are 15\% Human3.6m~\cite{h36m}, 25\% COCO~\cite{coco}, 5\% 3DPW~\cite{3dpw}, 30\% AGORA~\cite{patel2021agora}, 25\% Model Agency Dataset~\cite{choutas2022accurate}. In each iteration, we also add 50\% synthetic data generated with the same setting as~\cite{sengupta2021hierarchical}.

\begin{figure}[b]
	\centering
	\vspace{-18pt}
	\includegraphics[width=0.7\linewidth]{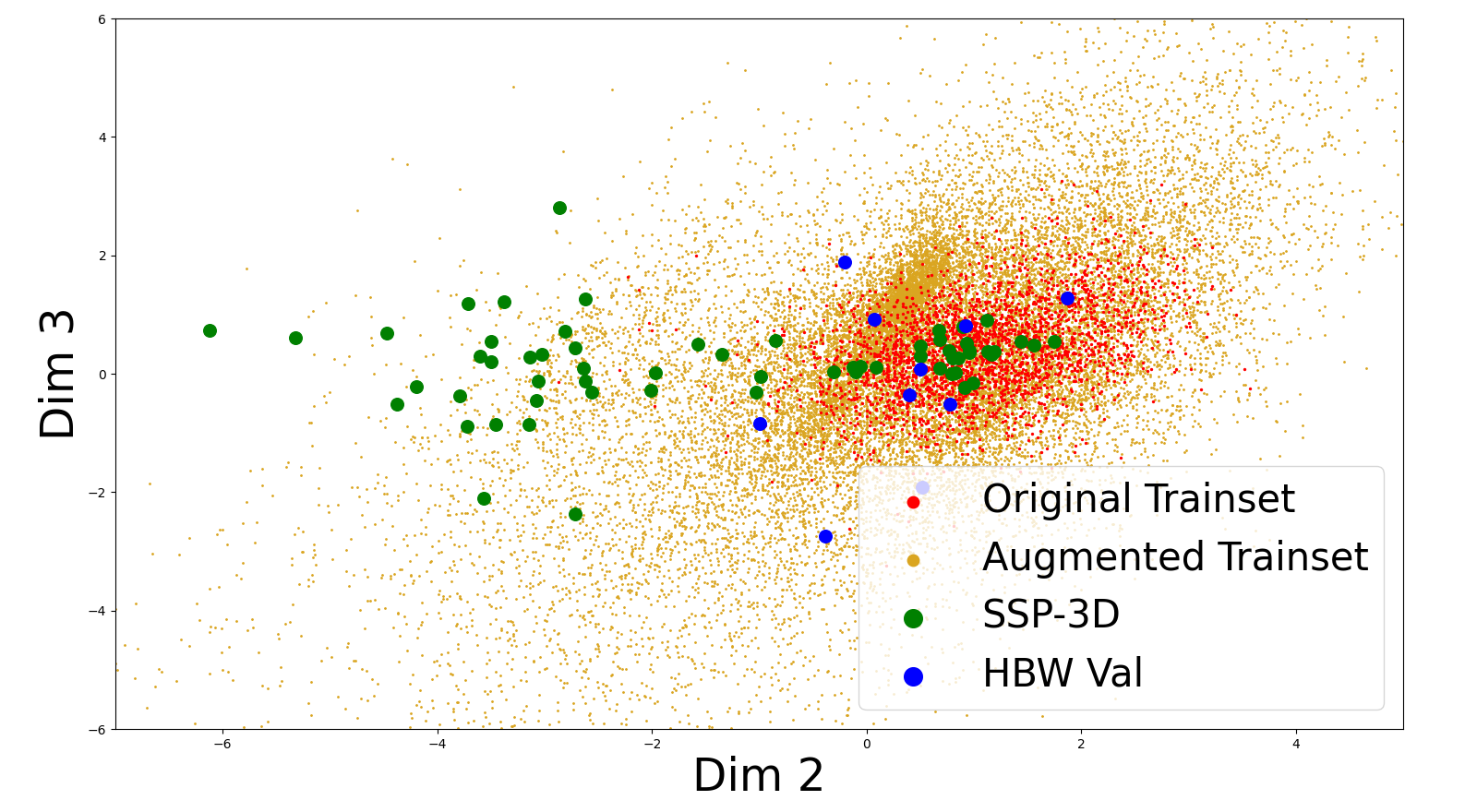}
	\vspace{-7pt}  
	\caption{2nd and 3rd shape coefficient in different datasets.}
	\vspace{-5pt} 
\end{figure}
\begin{figure}[t]
  \centering
    \begin{subfigure}{.23\textwidth}
      \centering
      \includegraphics[width=\linewidth]{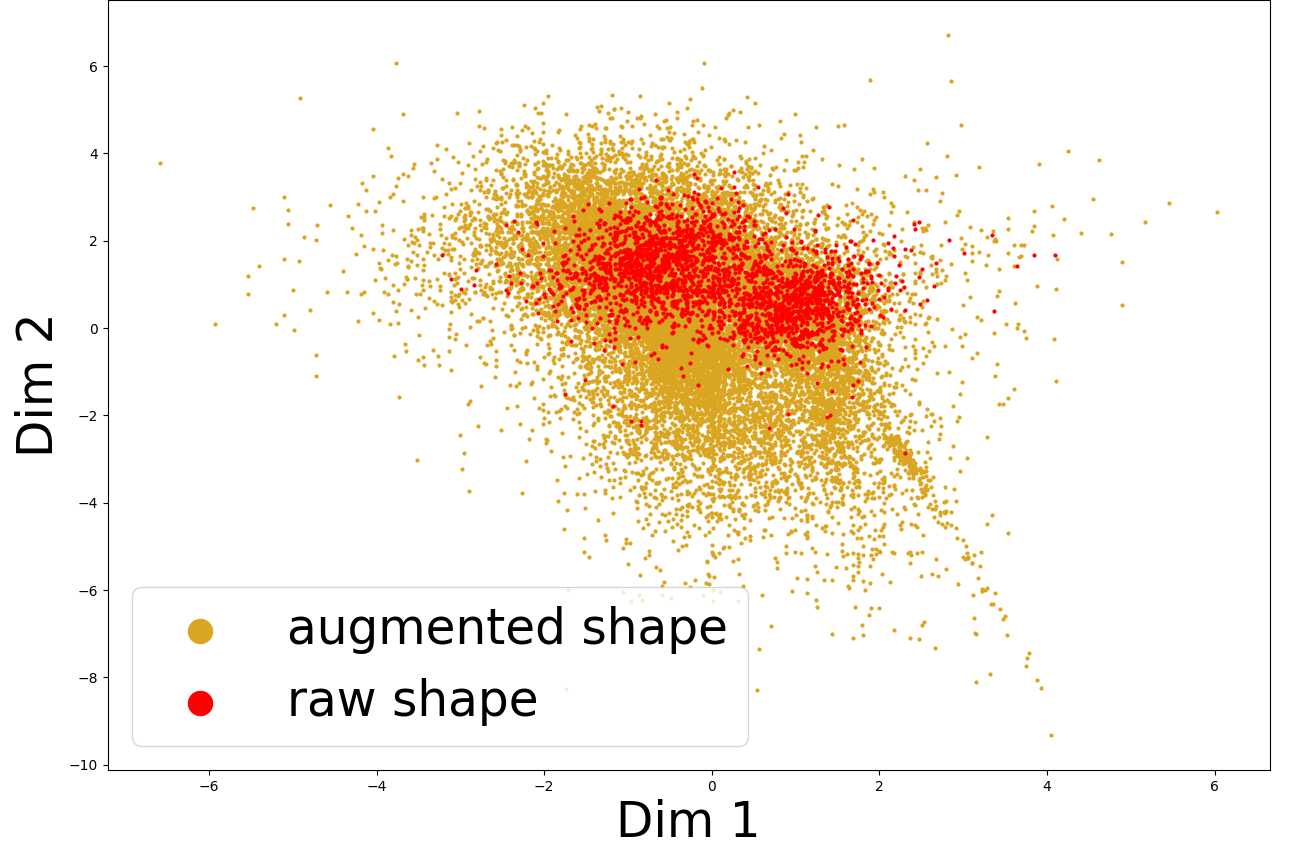}  
      \caption{First two dim. of shape PCA}
      \label{fig:div1} 
    \end{subfigure}
    \begin{subfigure}{.23\textwidth}
      \centering
      \includegraphics[width=\linewidth]{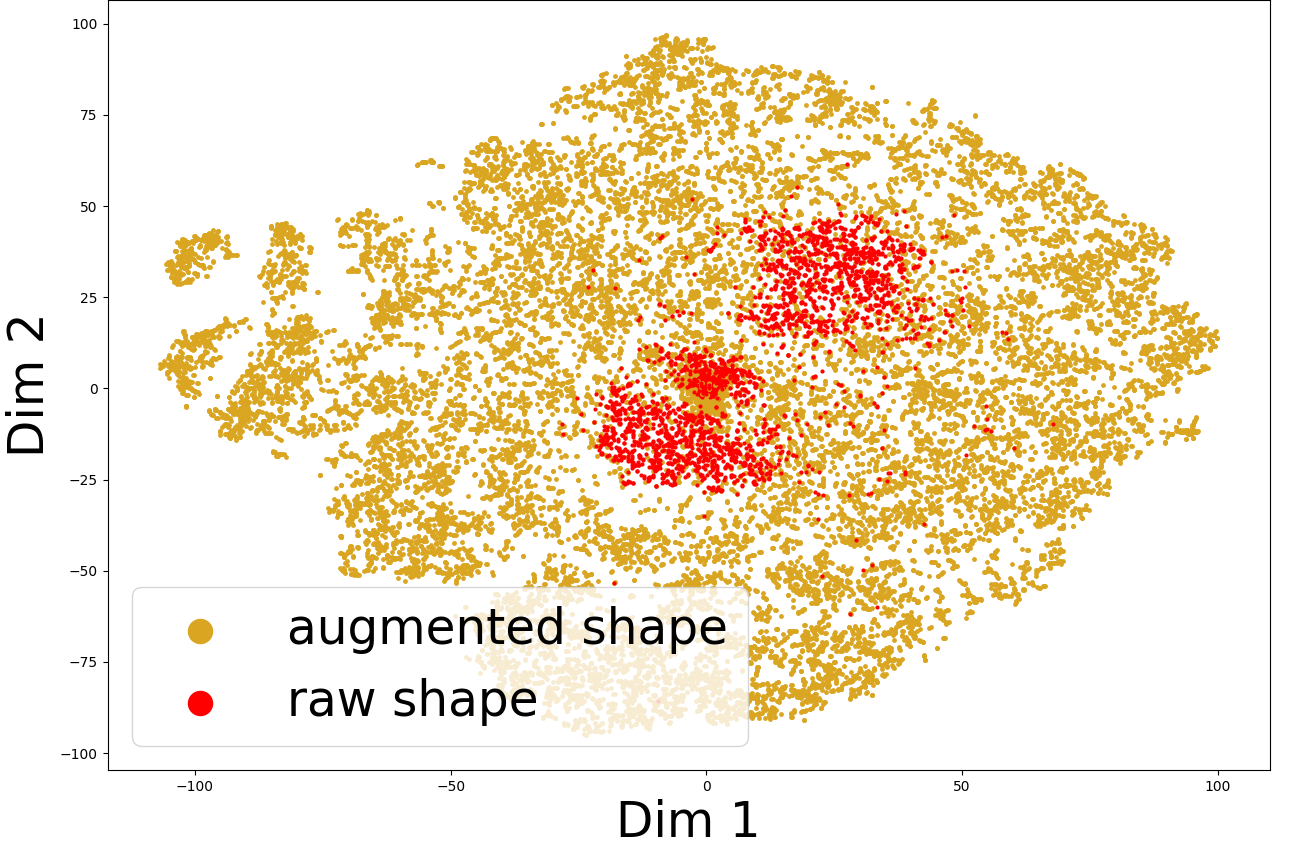}  
      \caption{t-SNE result of shape.}
      \label{fig:div2}
    \end{subfigure}
    \label{fig:div}
    \caption{Visualization of human shape distribution before and after augmentation.}
\end{figure}

We evaluate our method on HBW and SSP-3D datasets. In Fig. \ref{fig:div1}, we visualize the 2nd and 3rd shape coefficients of $\beta$ in different datasets. It shows the body shapes in the augmented training set and SSP-3D are more diverse than other datasets. On the contrary, the diversity of clothing in HBW dataset surpasses that in SSP-3D dataset. SSP-3D mainly contains people in tight or minimal clothing, whereas HBW dataset has a broader range of clothing types, including T-shirts, sweaters, dresses, thick jackets, etc.

\paragraph{Training and Loss} Our model undergoes 80000 iterations of training with the Adam solver. The learning rate is initially set to $1\times10^{-3}$ at first and decreased by a factor of 10 after 60000 iterations. The training is performed with a mini-batch size of 32 per GPU and utilizes 4 GPUs in total. Implementation is in PyTorch.  

The scalar coefficients in the loss function are $\mu_0=0.01$, $\mu_1=0.01$, $\mu_2=0.1$, $\mu_3=1$. 
\section{Extra Experiments}\label{sec:exp}

\subsection{Pose Estimation Experiments}

\begin{table*}[tbh]
	\begin{center}
    {
        \begin{tabular}{l|l|cc}
            \toprule 
            & Model & MPJPE & PA-MPJPE \\
            \midrule 
            HMR~\cite{hmr} & SMPL & 130.0 & 81.3 \\
            SPIN~\cite{spin} & SMPL & 96.9 & 59.2 \\
            \cite{sengupta2020synthetic} & SMPL & - & 66.8 \\
            ExPose~\cite{choutas2020monocular} & SMPL-X & 93.4 & 60.7 \\
            EFT~\cite{joo2021exemplar} & SMPL & 85.1 & 52.2 \\
            \cite{sengupta2021hierarchical} & SMPL & 84.9 & 53.6 \\
            HybrIK~\cite{li2021hybrik} & SMPL & 80.0 & 48.8 \\
            PARE~\cite{kocabas2021pare} & SMPL & \textbf{74.5} & 46.5 \\
            SHAPY~\cite{choutas2022accurate} & SMPL-X & 95.2 & 62.6 \\
            \midrule
            ShapeBoost (Ours) & SMPL & 75.3 & \textbf{44.6}\\
            \bottomrule
        \end{tabular}
    }
\caption{\textbf{Quantitative comparisons for pose estimation} on 3DPW~\cite{3dpw}. }
\label{table:supp_pose_3dpw}
\end{center}
\end{table*}{} 

We compare our results of pose estimation with previous methods on 3DPW dataset~\cite{3dpw}. Since our method uses the pose estimation backbone of HybrIK~\cite{li2021hybrik}, we also retrain HybrIK~\cite{li2021hybrik} using the same backbone and training datasets as our method for a fair comparison. The results are shown in Tab.~\ref{table:supp_pose_3dpw}. 

From Tab.~\ref{table:supp_pose_3dpw}, we can find that our model achieves more accurate pose estimation results than previous methods. The pose estimation score shows that our method achieves pixel-level alignment to the input images. Compared to previous methods that directly predict the shape parameter from the image, our model first predicts the joint coordinates, and then recovers the shape based on the extracted skeleton. As a result, the resulting shape is more consistent with the keypoint predictions and shows better image alignment.

\subsection{Analysis of Data Augmentation}
\textcolor{black}{To demonstrate the effectiveness of our data augmentation module, we first visualize the shape distribution before and after augmentation in Fig.\ref{fig:div1} \ref{fig:div2}. In Fig. \ref{fig:div1}, we show the first 2 dimensions of shape PCA, and in Fig. \ref{fig:div2}, we show the t-SNE result of body shapes. We can find that the shape distribution after augmentation distributes covers the old distribution and also covers more area. This shows that our data augmentation module can greatly increase the diversity of the training data.}

\subsection{Influence of Pose-dependent Shape Deformation}
In our experiments, we always predict bone lengths and part widths in rest pose SMPL model instead of the values in the posed SMPL model, and do not take into account the pose-dependent shape deformation. According to our experiments on AMASS dataset, pose-dependent deformations influences 0.1\% in bone lengths and 2\% in part widths in average, which are minor.

\section{Converting SMPL Prediction to SMPL-X}
\label{sec:smpl2smplx}
The proposed dataset labels and pretrained models in SHAPY~\cite{choutas2022accurate} all use SMPL-X~\cite{pavlakos2019expressive} model. Since the shape space of SMPL and SMPL-X model are different, our SMPL-based prediction suffers from systematic error. Therefore, we convert the predicted SMPL mesh to SMPL-X using the least squares method based on the point regressor provided in SHAPY~\cite{choutas2022accurate}.

SHAPY~\cite{choutas2022accurate} uniformly samples the SMPL-X template mesh and proposes a sparse matrix $\mathbf{H}_{\textrm{SMPL-X}} \in \mathbb{R}^{p \times N}$ to regress the sampled points from SMPL-X vertices $\mathbf{T}_\textrm{SMPL-X}$, as $\mathbf{P}=\mathbf{H}_{\textrm{SMPL-X}} \mathbf{T}_\textrm{SMPL-X}$, where $p$ is the sampling number, and $N$ is the vertex number of SMPL-X model. Then they register the same set of points on SMPL model and compute $\mathbf{H}_{\textrm{SMPL}} \in \mathbb{R}^{p \times K}$, where $K$ is the vertex number of SMPL model.

Given a rest-pose SMPL mesh $\mathbf{T}_\textrm{SMPL}$, our goal is to find a shape parameter for SMPL-X model, denoted as $\boldsymbol{\beta}_\textrm{SMPL-X}$ so that the regressed mesh surface points are best aligned. The problem can be written as a linear L2-norm approximation problem:
\begin{align}
    &\qquad\quad \boldsymbol{\beta}_\textrm{SMPL-X}, \mathbf{t}_\textrm{SMPL-X} = \argmin_{\boldsymbol{\beta}, \mathbf{t}} \left\| \Delta \mathbf{P} \right\|_2^2 \\
    &\mathrm{where, \ } \left\{ 
        \begin{aligned} 
        \Delta \mathbf{P} &= \mathbf{H}_{\textrm{SMPL-X}} \mathbf{T} - \mathbf{H}_{\textrm{SMPL}} \mathbf{T}_\textrm{SMPL} - \mathbf{t} \\
        \mathbf{T} &= \mathbf{S}_\textrm{SMPL-X} \boldsymbol{\beta} + \mathbf{T}^0_\textrm{SMPL-X}
    \end{aligned} 
    \right. .
\end{align}


where $\mathbf{t}\in \mathbb{R}^3$ is the global translation vector, $\mathbf{T}^0_\textrm{SMPL-X}\in \mathbb{R}^{N\times 3}$ is the template mesh of SMPL-X model, and $ \mathbf{S}_\textrm{SMPL-X}\in \mathbb{R}^{3N\times s}$ is the weight matrix that constructs the rest-pose SMPL-X mesh from shape parameters. $s$ is the dimension of the shape parameter used in SMPL-X. In our implementation, we use $s=10$. This linear optimization problem can be solved analytically.

To simplify the equation, we denote
\begin{equation}
    \mathbf{E} = \left[
        \begin{array}{ccc}
            -1 & 0 & 0 \\
            0 & -1 & 0 \\
            0 & 0 & -1 \\
            -1 & 0 & 0 \\
            0 & -1 & 0 \\
            \vdots & \vdots & \vdots \\
            0 & 0 & -1 \\
        \end{array}
    \right] \in \mathbb{R}^{3p \times 3},
\end{equation}
\begin{align}
    \mathbf{A} &= \left[
        \begin{array}{cc}
            \mathbf{H}_{\textrm{SMPL-X}}\mathbf{S}_\textrm{SMPL-X} & \mathbf{E} \\
        \end{array}
    \right] \in \mathbb{R}^{3p\times(s+3)},\\
    \mathbf{b} &= \mathbf{H}_{\textrm{SMPL}}\mathbf{T}_\textrm{SMPL} - \mathbf{H}_{\textrm{SMPL-X}}\mathbf{T}^0_\textrm{SMPL-X} \in \mathbb{R}^{3p}.
\end{align}
Then the optimization problem is transformed into finding the least squares solution of the overdetermined linear system: 
\begin{equation}
    \mathbf{A} \left[
        \begin{array}{c}
            \boldsymbol{\beta} \\
            \mathbf{t} \\
        \end{array}
    \right] = \mathbf{b}.
\end{equation}
The solution is:
\begin{equation}
    \left[
        \begin{array}{c}
            \boldsymbol{\beta}_\textrm{SMPL-X} \\
            \mathbf{t}_\textrm{SMPL-X} \\
        \end{array}
    \right] = (\mathbf{A} ^T\mathbf{A} )^{-1}\mathbf{A} ^T\mathbf{b}.
\end{equation}

Compared to the fitting method for model conversion, our method is much faster and not affected by the initialization of optimization.

\section{Limitations and Future Work}
\label{sec:limitation}
\begin{figure*}[ht]
    \centering
    \includegraphics[width=0.8\linewidth]{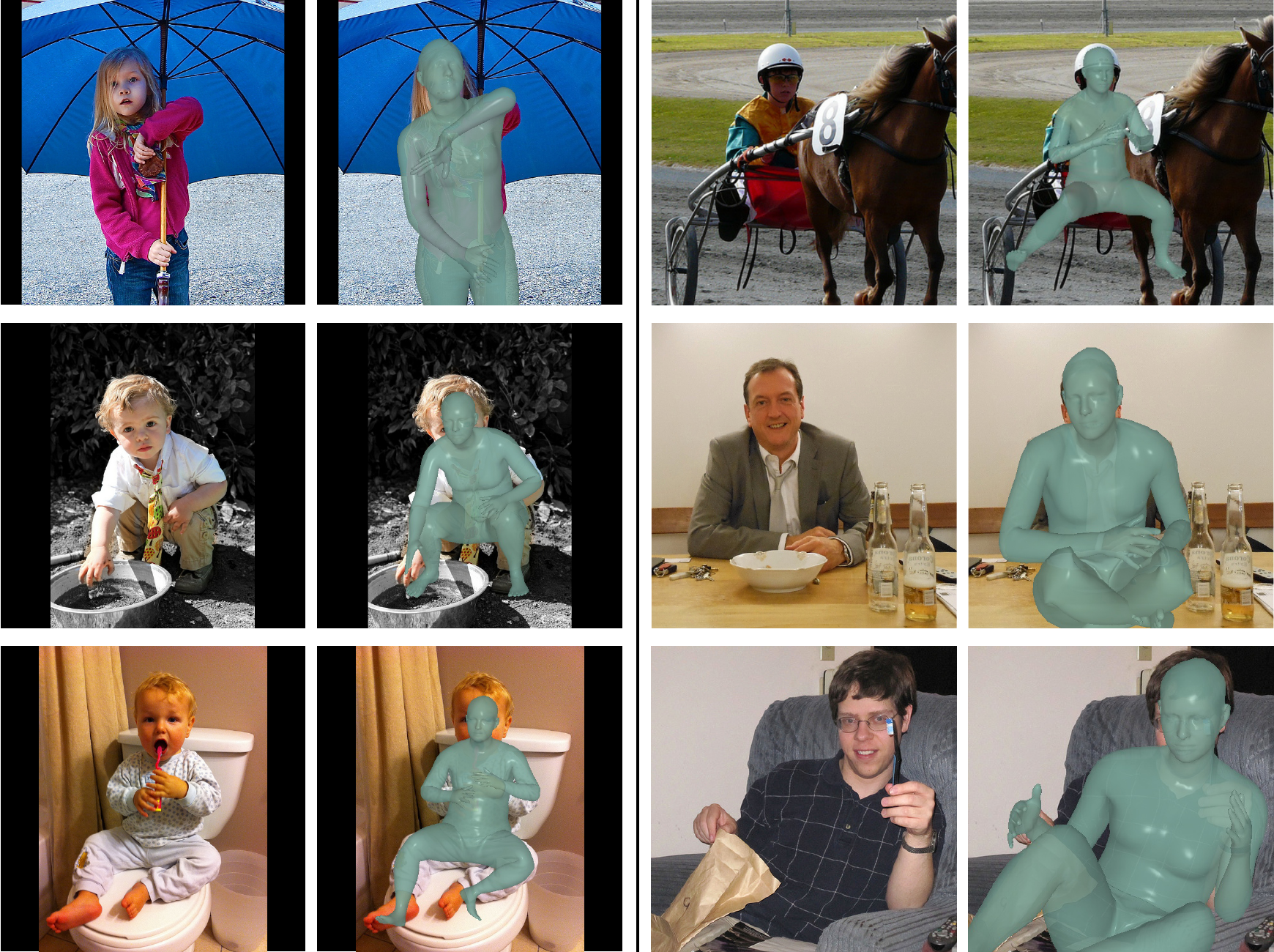}
    \caption{{Failure cases of ShapeBoost.} As shown in the left 2 columns, the diversity of ShapeBoost output is limited by the expressiveness of SMPL, making it hard to model the body shapes of children. As shown in the right 2 columns, ShapeBoost sometimes fails to give an accurate pose estimation in severe occlusion situations.}
    \label{fig:failure} 
\end{figure*}

Our work has several limitations as shown in Fig.~\ref{fig:failure}. First, ShapeBoost sometimes fails to give an accurate pose estimation in severe occlusion situations. Second, since ShapeBoost utilizes SMPL model, the diversity of the output shape is limited by the expressiveness of SMPL. For example, ShapeBoost does not provide accurate shape estimation for children or people with a muscular body. The first limitation can be mitigated by using a more robust pose estimation algorithm. The second limitation can be alleviated by using other body models such as SMIL~\cite{hesse2018learning} and STAR~\cite{osman2020star}.

\section{Qualitative Results}
\label{sec:vis}

\begin{figure*}[ht]
    \centering
    \includegraphics[width=\linewidth]{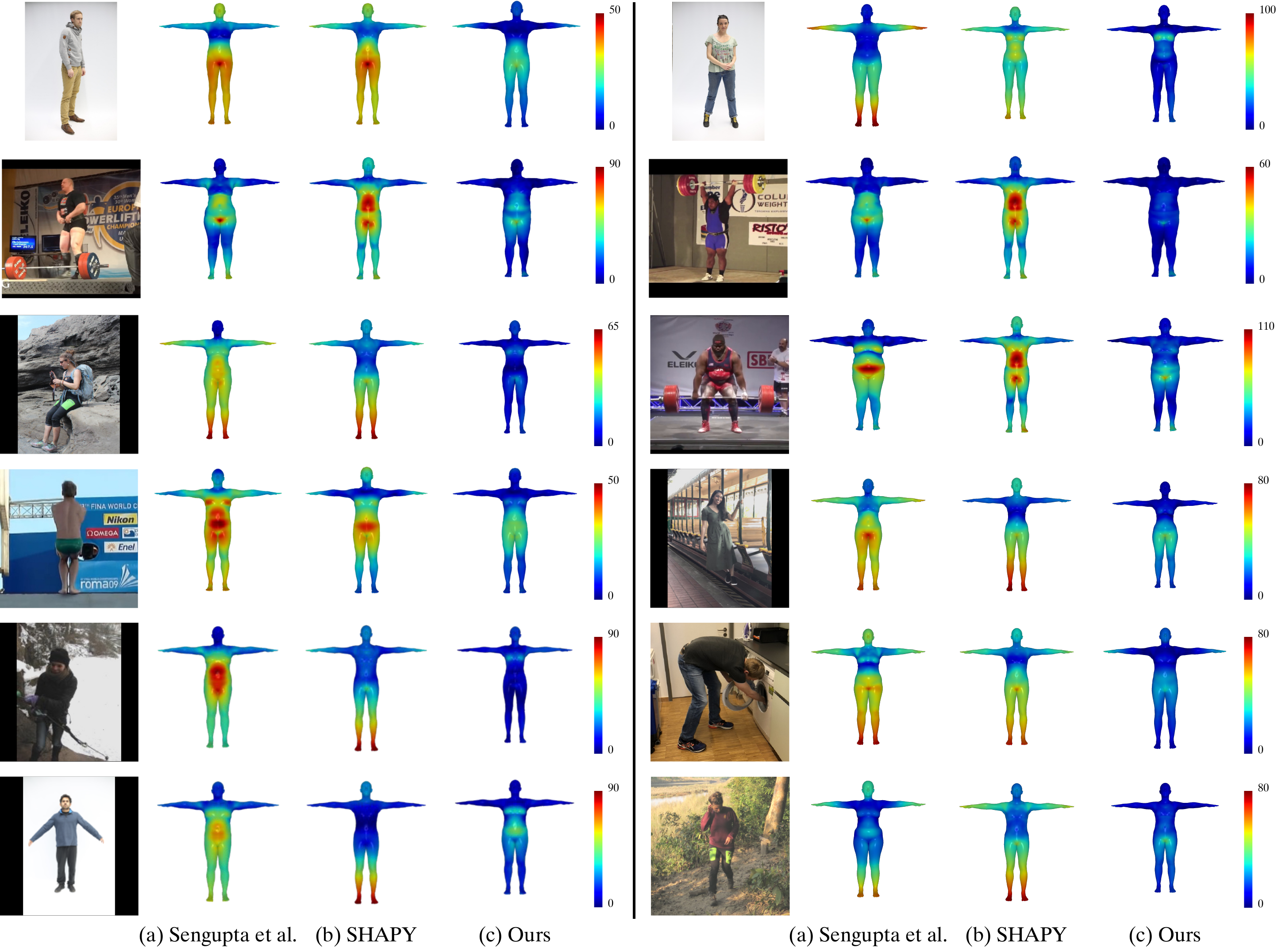}
    \caption{{Qualitative results on SSP-3D~\cite{sengupta2020synthetic} and HBW~\cite{choutas2022accurate} datasets.} From left to right: Input image, (a) Sengupta et al.~\cite{sengupta2021hierarchical} results, (b) SHAPY~\cite{choutas2022accurate} results, and (c) Our results. Warmer colors mean higher per-vertex error. Experiments on SSP-3D dataset use PVE-T-SC metric, and experiments on HBW dataset use P2P$_{20\mathrm{K}}$ metric. Unit: mm.}
    \label{fig:compare_vshaped} 
\end{figure*}
\begin{figure*}[t]
    \begin{subfigure}{\linewidth}
      \centering
      \includegraphics[width=\linewidth]{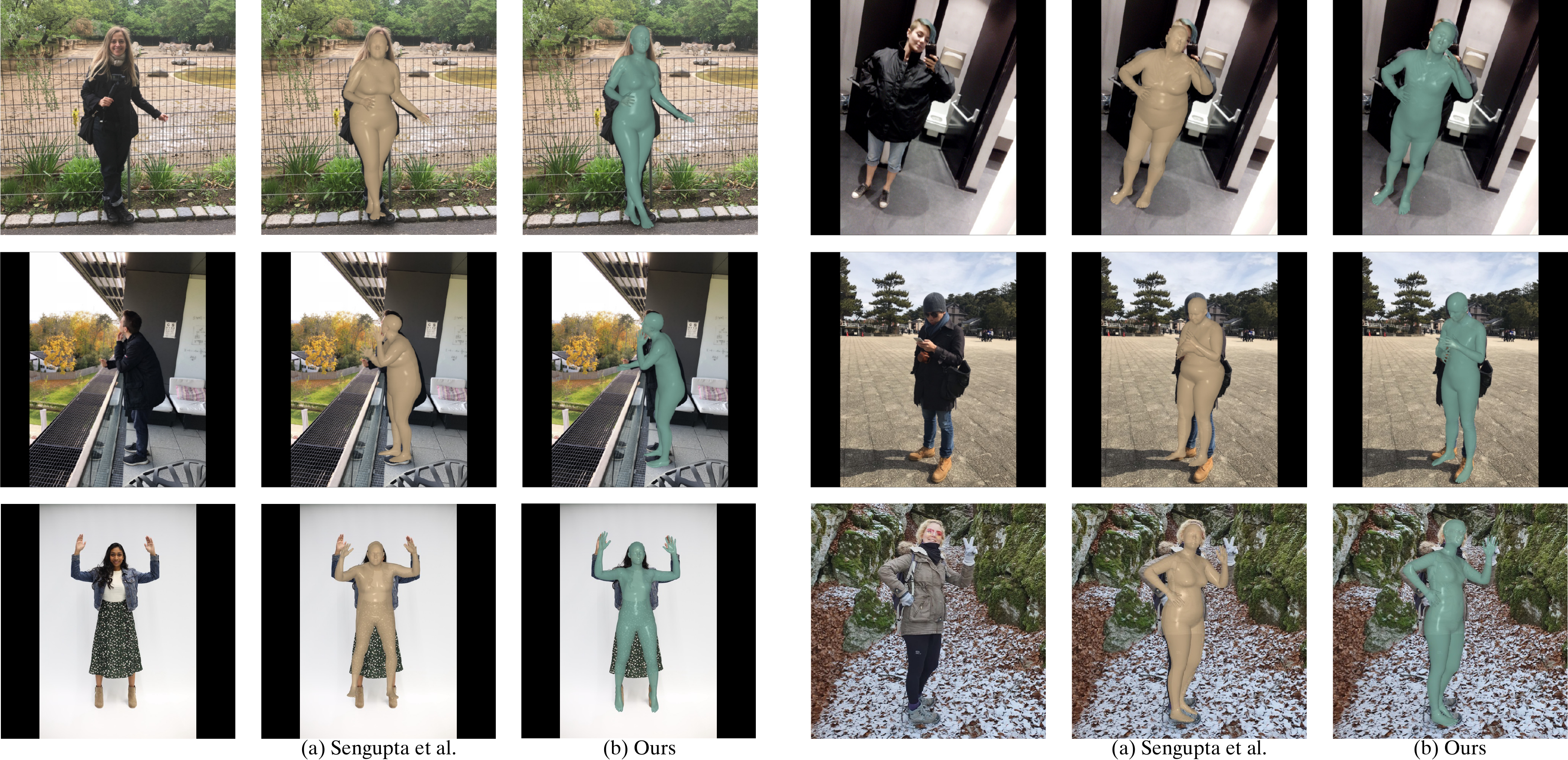} 
        \caption{{Qualitative comparison between Sengupta et al.~\cite{sengupta2021hierarchical} and ShapeBoost (Ours).} From left to right: Input image, (a) Sengupta et al.~\cite{choutas2022accurate} results, and (b) Our results. Our method is better aligned to the input especially for images of people in thick clothes.\\}
        \label{fig:compare_hie} 
    \end{subfigure}
    \begin{subfigure}{\linewidth}
      \centering
      \includegraphics[width=\linewidth]{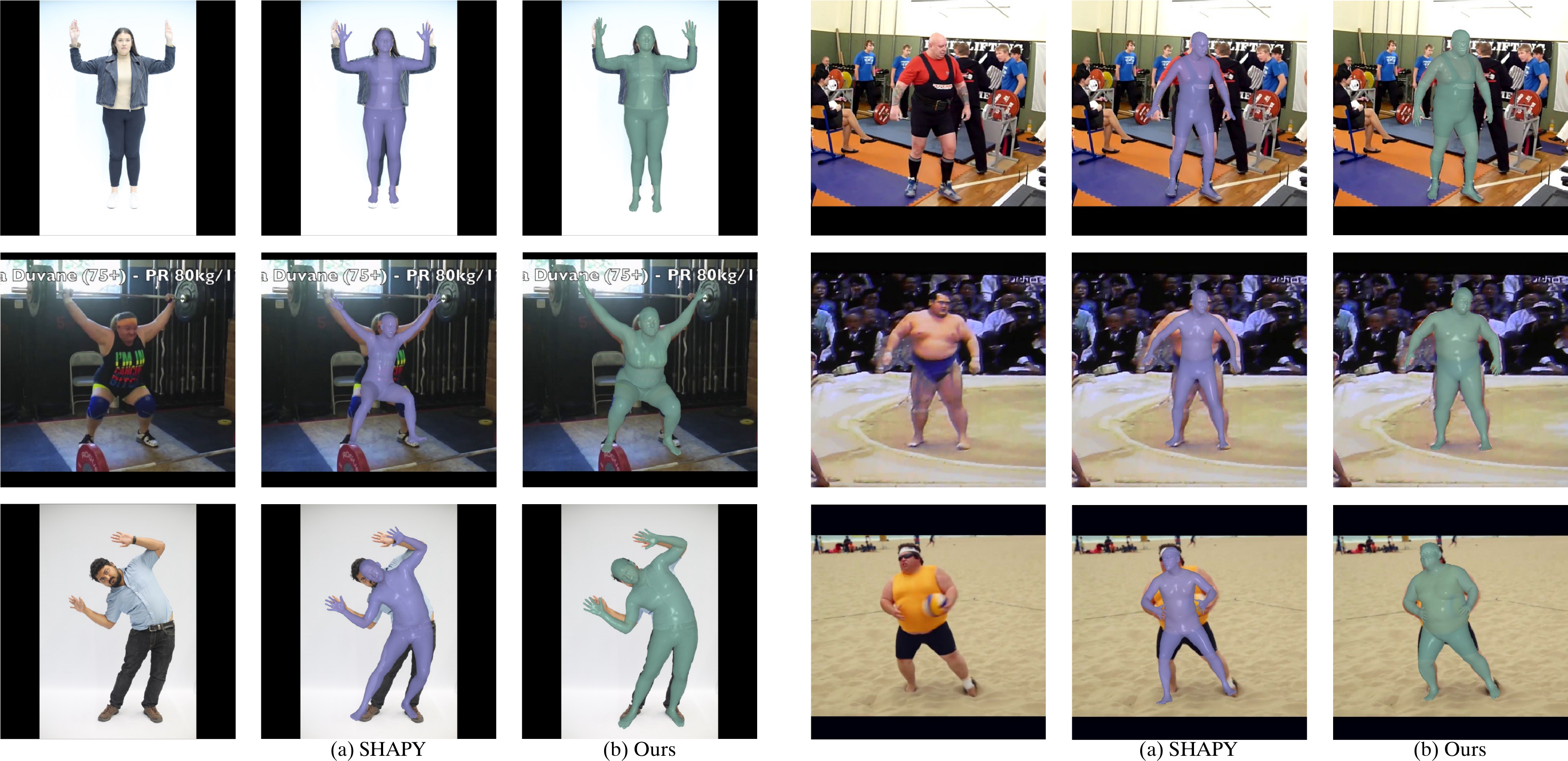} 
        \caption{{Qualitative comparison between SHAPY~\cite{choutas2022accurate} and ShapeBoost (Ours).} From left to right: Input image, (a) SHAPY~\cite{choutas2022accurate} results, and (b) Our results. Our method is better aligned to the input especially for images of chubby people.}
        \label{fig:compare_shapy} 
    \end{subfigure}
    \caption{Qualitative comparison with previous methods.}
\end{figure*}
\begin{figure*}[b]
    \centering
    \includegraphics[width=\linewidth]{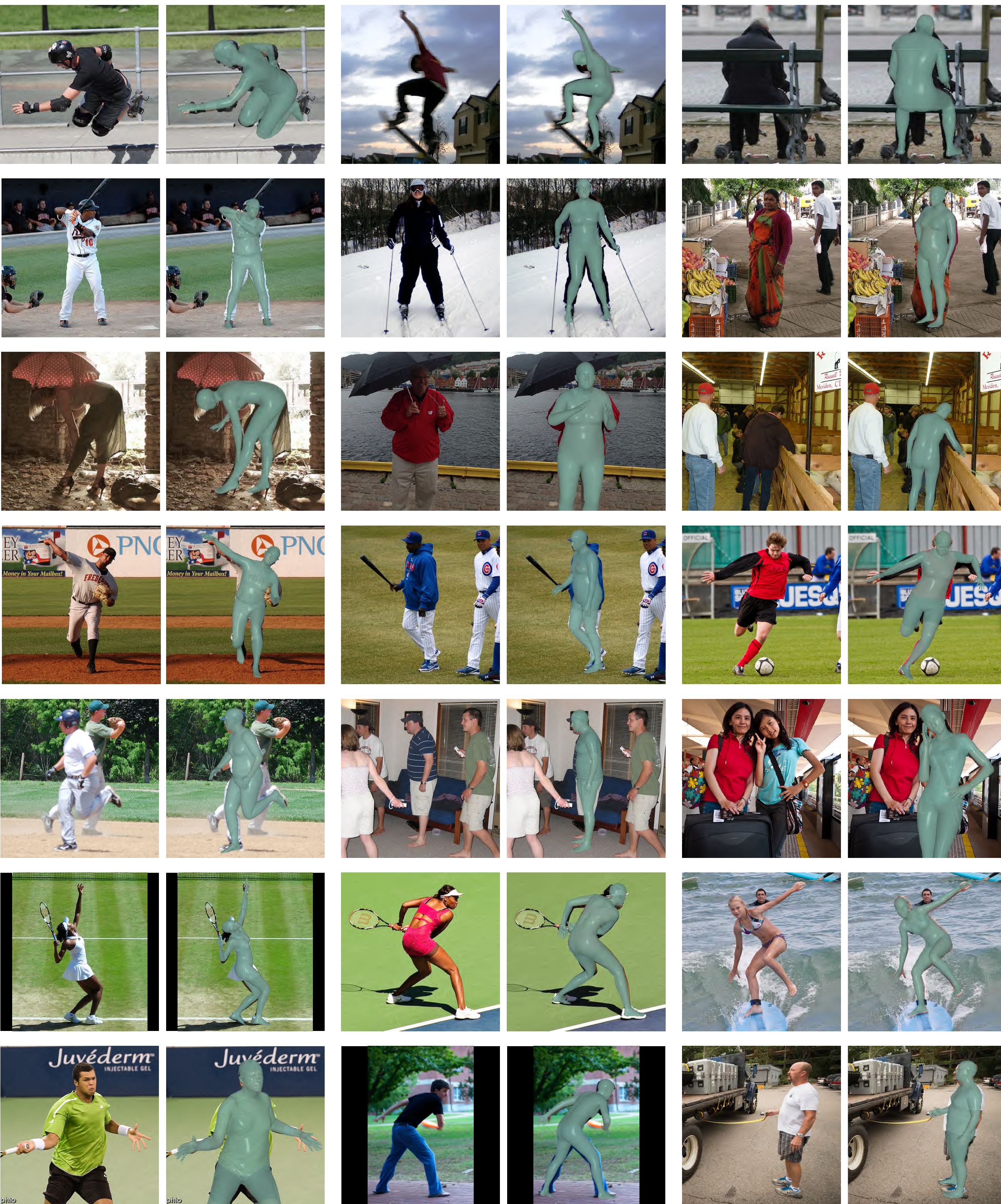} 
    \caption{{Qualitative results on COCO~\cite{coco} dataset.} Our methods predicts accurate results for images of people with hard pose, in occlusion, wearing loose clothes, and with an extreme body shape.}
    \label{fig:wild} 
\end{figure*}

We provide additional qualitative results. In Fig.~\ref{fig:compare_vshaped}, we compare the results predicted by different models, and visualize their per-vertex error using the heatmap. In Fig.~\ref{fig:compare_hie}, we compare the results of Sengupta et al.~\cite{sengupta2021hierarchical} and ShapeBoost, and in Fig.~\ref{fig:compare_shapy}, we compare the results of SHAPY~\cite{choutas2022accurate} and ShapeBoost. The method proposed by Sengupta et al. often fails on images with people in thick clothes, and SHAPY often fails on extreme body shapes. In comparison, our method is accurately aligned to the input image in different scenarios. In Fig.~\ref{fig:wild}, we visualize the results of our model on in-the-wild images. The results show that ShapeBoost can handle images with hard pose, thick clothes and extreme body shapes.
~\newpage

\bibliography{aaai24} 

\end{document}